\documentclass[%
 aip,
 amsmath,amssymb,
preprint,
]{revtex4-2}
\usepackage[cal=boondox]{mathalfa}
\usepackage{graphicx}% Include figure files
\usepackage{dcolumn}% Align table columns on decimal point
\usepackage{bm}% bold math
\usepackage{tabularx}
\usepackage{array}     % 提供 \arraybackslash
\newcolumntype{C}{>{\centering\arraybackslash}X} % 置中可伸縮欄位
\usepackage{xcolor}
\usepackage[utf8]{inputenc}
\usepackage[T1]{fontenc}
\usepackage{mathptmx}
\usepackage{etoolbox}
\usepackage{caption}
\usepackage{subcaption}
\usepackage{multirow}
\usepackage[ruled]{algorithm2e}
\usepackage[hidelinks]{hyperref}
\usepackage{amsmath}   % 你表格裡有 \tfrac、\text，需要它
\usepackage{makecell}  % 讓單一儲存格換行

\makeatletter
\def\@email#1#2{%
  \endgroup
  \patchcmd{\titleblock@produce}
    {\frontmatter@RRAPformat}
    {\frontmatter@RRAPformat{\produce@RRAP{*#1\protect\href{mailto:#2}{#2}}}}%
    {}{}%
}
\makeatother

\makeatletter
\let\orig@maketitle\maketitle
\renewcommand{\maketitle}{%
  \begingroup
    \let\raggedright\centering % 把內部用到的 \raggedright 暫時改成 \centering
    \orig@maketitle            % 呼叫原本的 revtex \maketitle（會輸出 abstract）
  \endgroup
}
\makeatother
\begin{document}

%\reprint{aa}

\title{Impact of Loss Weight and Model Complexity on Physics-Informed Neural Networks
for Computational Fluid Dynamics}

\author{Yi-En Chou$^{1}$}
\author{Te-Hsin Liu$^{1}$}
\author{Chao-An Lin$^{1}$}
\affiliation{$^{1}$ Department of Power Mechanical Engineering,
National Tsing Hua University, Hsinchu 30013, Taiwan \\
Email address: \\ dodger25685@gmail.com (Y.-E. Chou), \\ hsinl606@gmail.com (T.-H. Liu), \\calin@pme.nthu.edu.tw (C.-A. Lin)}

% \title{\begin{center}
% Impact of Loss Weight and Model Complexity on Physics-Informed Neural Networks for Computational Fluid Dynamics
% \end{center}}
% \author{
%     \centering
%     Yi-En Chou$^1$, Chao-An Lin$^{1,*}$ \\
%     $^1$Department of Power Mechanical Engineering, National Tsing Hua University, \\ Hsinchu 30013, Taiwan \\
%     \textit{\footnotesize *Corresponding author: calin@pme.nthu.edu.tw (C.A. Lin)}
% }

%\date{\today}

%%%%%%%%%%%%%%%%%%%%%%%%%%%%%%%%%%%%%%%%%%%%%%%%%%%%%%%
%             ABSTRACT
%%%%%%%%%%%%%%%%%%%%%%%%%%%%%%%%%%%%%%%%%%%%%%%%%%%%%%%

\begin{abstract}
Physics-Informed Neural Networks (PINNs) offer a mesh-free framework for solving PDEs but are highly sensitive to loss weight selection. We propose two dimensional-analysis-based weighting schemes: one based on quantifiable terms, and another also incorporating unquantifiable terms for more balanced training. Benchmarks on heat conduction, convection–diffusion, and lid-driven cavity flows show that the second scheme consistently improves stability and accuracy over equal weighting. Notably, in high-Peclet-number convection–diffusion, where traditional solvers fail, PINNs with our scheme achieve stable, accurate predictions, highlighting their robustness and generalizability in CFD problems.
\end{abstract}

\maketitle

%%%%%%%%%%%%%%%%%%%%%%%%%%%%%%%%%%%%%%%%%%%%%%%%%%%%%%%
%             INTRODUCTION
%%%%%%%%%%%%%%%%%%%%%%%%%%%%%%%%%%%%%%%%%%%%%%%%%%%%%%%

\section{Introduction}\label{sec:level1}

In this study, we apply deep learning based method to computational fluid dynamics(CFD). In this chapter, we will begin from introducing some background knowledge of deep learning which is correlated to this work, followed by literature survey, and last, we will introduce the organization of this thesis in summary.

\subsection{Deep learning}\label{sec1.1}
\subsubsection{Artificial Intelligence(AI), Machine Learning(ML) and
Deep Learning(DL)}

\begin{figure}[htbp!]
    \centering
    \includegraphics[width=0.6\textwidth]{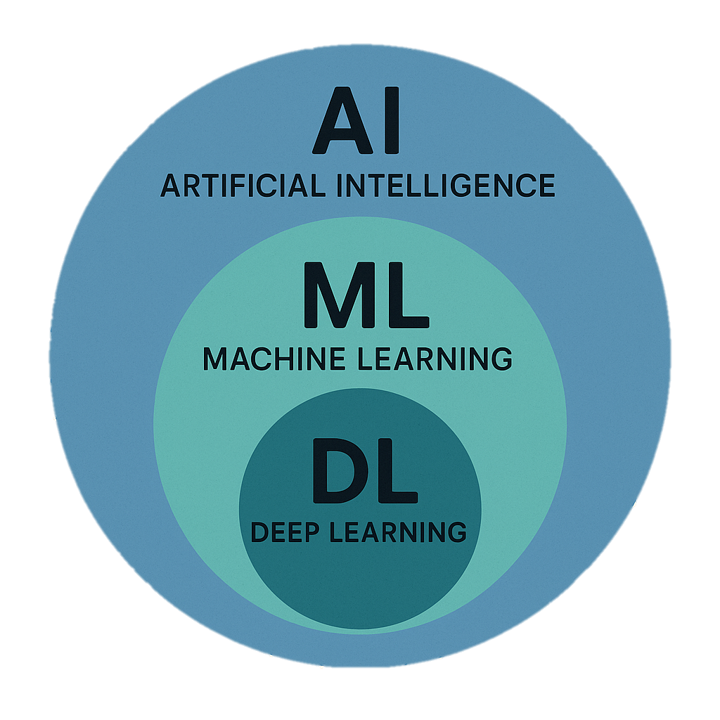}
    \caption{Relationship between AI, ML, and DL.}
\end{figure}

Artificial Intelligence (AI) refers to computational techniques that enable machines to mimic human intelligence. It encompasses a wide range of methods, including symbolic reasoning \cite{Kolata1982}, expert systems \cite{Jackson1986}, natural language processing \cite{Chowdhary2020}, and more recently, machine learning (ML) \cite{Jordan2015} and deep learning (DL) \cite{Lecun2015}.
ML focuses on algorithms that automatically learn from data without explicit programming. As noted by Tom M. Mitchell \cite{Jordan2015}, “A computer program is said to learn from experience E with respect to some class of tasks T and performance measure P, if its performance at tasks in T, as measured by P, improves with experience E.”

\subsubsection{Deep neural networks}
A deep neural network (DNN)~\cite{Goodfellow2016} is inspired by the biological neural network of the brain. In biology, neurons sum incoming signals and fire when a threshold of excitation is reached. Artificial neurons mimic this behavior using weighted sums and activation functions. The first artificial neuron, the perceptron, was introduced by Rosenblatt in 1958~\cite{Rosenblatt1958} as a linear classifier that categorized inputs into two possible classes.  

A neural network consists of an input layer $\mathbf{x}$, multiple hidden layers $\mathbf{h}_{1}, \mathbf{h}_{2}, \dots, \mathbf{h}_{H}$, and an output layer. Each neuron in the hidden layer computes a weighted sum of its inputs with parameters $\{W_i^T, b_i\}$, then applies a nonlinear activation function $\phi$:  
\begin{equation}
\mathbf{h}_{1} = \phi \big(W_0^T \cdot \mathbf{x}_i + b_0 \big).
\end{equation}
Activation functions play a crucial role in neural networks by introducing nonlinearity and enabling the network to capture complex relationships between inputs and outputs. Without nonlinear activation, even very deep neural networks would behave as linear models; with nonlinear activation, they are able to tackle complex tasks.  

Training a neural network can be summarized into the following steps: (i) a forward pass, where the input data is propagated through the network to generate predictions; (ii) computation of the loss $\mathcal{L}$, which measures the difference between predictions and ground truth, followed by a backward pass to determine parameter gradients; and (iii) parameter updates based on these gradients. This process is repeated for many iterations (commonly called \emph{epochs}) until convergence. The model parameters are defined as  
\begin{equation}
\theta = \{\omega, b\},
\end{equation}
where $\omega$ are the weights and $b$ the biases.  

Different tasks often favor different architectures: convolutional neural networks (CNNs) for vision, recurrent neural networks (RNNs) for audio, and generative models such as VAEs~\cite{Cinelli2021} and GANs~\cite{Goodfellow2014} for data generation. In contrast, Physics-Informed Neural Networks (PINNs) typically adopt fully-connected networks (FCNs)~\cite{Schwing2015}, which require no special assumptions about the input and therefore provide a flexible starting point with wide applicability.  

\subsection{Literature survey}\label{sec1.2}
\subsubsection{Machine Learning for Computational Fluid Dynamics}
Although the origins of machine learning date back to the 1940s, its application to computational fluid dynamics (CFD) has emerged only in recent years. Brunton et al. \cite{Vinuesa2022} highlighted the potential of integrating machine learning with CFD, outlining three main directions: accelerating Direct Numerical Simulation (DNS), improving turbulence modeling, and developing Reduced-Order Models (ROMs).

\paragraph{Increasing the speed of DNS}:

Turbulence can be simulated using RANS, LES, or DNS. Among them, DNS provides the highest fidelity but is also the most computationally expensive, with costs rising sharply as Reynolds number increases. To improve efficiency, researchers have proposed methods such as reducing resolution requirements \cite{Hickey2019}, accelerating the solution of the Poisson equation \cite{Hickey2019,Sprechmann2022}, and other techniques.

\paragraph{Modelling improvement}:

RANS offers the highest computational efficiency but the lowest accuracy. Conventional models are typically based on the Boussinesq approximation, relying on extensive mathematical derivations and assumptions \cite{Huang1998}. Beyond these approaches, recent studies have applied machine learning to enhance RANS models, including improving numerical stability \cite{Mcconkey2022} and predicting Reynolds stresses with physics-informed neural networks \cite{Kurzawski2016}, among others.

\paragraph{Reduced-Order modelling(ROM)}:

Reduced-Order Modeling (ROM) seeks low-dimensional representations of flow fields, reducing both memory requirements and computational cost—an important advantage given GPGPU memory limits and the strong dependence of simulation time on mesh size. ROM exploits the fact that even complex flows often exhibit dominant coherent structures \cite{Vinuesa2022}. A well-constructed representation not only improves computational efficiency but also enables accurate feature capture and serves as a tool for flow control \cite{Brunton2017}.

ROM has been applied in various fluid dynamics problems, including flow over a cylinder \cite{Morzynski2003}, incompressible flow over a flat-plate wing \cite{Brunton2017}, and turbulent flow over a NACA0012 airfoil \cite{Brunton2017}. Deep learning methods such as autoencoders further facilitate ROM by learning compact representations through a bottleneck architecture, where input data are encoded into a low-dimensional space and decoded back to approximate the original field \cite{Zhao2021,Agostini2020}.

\paragraph{Eulerian fluid simulation with neural network}:

Applications of neural networks to PDEs date back to the 1990s \cite{Lee1990}. To address the high cost of solving the incompressible Navier–Stokes equations, Schlachter et al. proposed a pressure prediction model using long-term convolutional neural networks to replace iterative solvers \cite{Sprechmann2022}. Other studies explored alternative approaches for solving the Poisson equation, including CNN-based solvers \cite{Wang2020,Hamzehloo2021}.

\subsubsection{Physics Inform Neural Networks(PINN)}

Physics-Informed Neural Networks (PINNs) \cite{Raissi01} combine deep learning with physics-based modeling by embedding governing equations into the training process. Beyond fitting data, PINNs incorporate loss terms that enforce physical laws, ensuring solutions remain consistent with both observations and underlying physics.
\begin{figure}[h]
    \centering
    \includegraphics[width=0.6\textwidth]{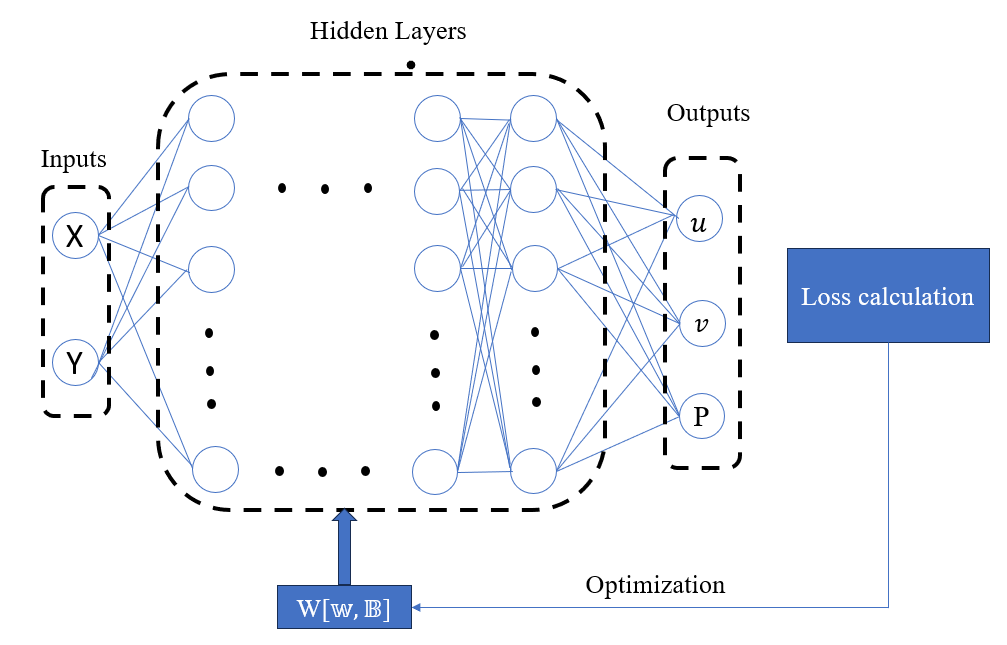}
    \caption{Schematic of training a physic-informed neural network (PINN).}
\end{figure}

\paragraph{Compute differential operators in PINN}:

Differentiation in PINNs is mainly computed by two approaches: automatic differentiation (AD) \cite{Jagtap2020,Bear2006} and numerical differentiation (ND) \cite{Ren2022}. AD, based on backpropagation and the chain rule, is widely used because it yields exact derivatives and aligns naturally with the meshless nature of PINNs. However, Chiu et al. \cite{Chiu2022} showed that AD may fail to capture physically consistent results since it does not correlate neighboring grid points. In contrast, ND correlates neighboring points and can sustain physical consistency, but introduces truncation errors.

\paragraph{Activation functions for PINN}:

Activation functions introduce nonlinearity, enabling neural networks to capture complex input–output relationships. Common choices in computer science include the sigmoid~\cite{Han1995} and ReLU~\cite{Glorot2011}, both valued for their simplicity and effectiveness~\cite{Lecun1998,Krizhevsky2017}. The sigmoid function,  

\begin{equation}
\phi(x) = \frac{1}{1 + e^{-x}},
\end{equation}

maps inputs to values between 0 and 1 and is widely used in binary classification tasks~\cite{Jeatrakul2009}. The ReLU function,  

\begin{equation}
\phi(x) = \max(0, x),
\end{equation}

is popular for overcoming the vanishing gradient problem and has proven effective in applications such as image recognition and natural language processing.  

In contrast, Physics-Informed Neural Networks (PINNs) often favor sinusoidal activations. PINNs solve partial differential equations (PDEs) by embedding physics-based constraints into neural networks, and studies have shown that the sine function is particularly well-suited for this purpose~\cite{Perdikaris2021,Weisstein2014}. The sine activation is defined as  

\begin{equation}
\phi(x) = \sin(x),
\end{equation}

and its periodic and smooth nature makes it effective for representing oscillatory or wave-like behaviors, naturally linking to generalized Fourier analysis.  

Although Fourier neural networks~\cite{Lapedes1987} are uncommon in general machine learning, several studies report that sinusoidal activations yield superior performance in PINNs~\cite{Raissi01,Sitzmann2020,Fang2021,Zobeiry2021} compared to ReLU or sigmoid. This advantage arises from the Fourier series property of the sine function, which ensures that any function can be represented as a series of sines~\cite{Perdikaris2021}, allowing PINNs to capture complex patterns, periodic phenomena, and sharp transitions with high accuracy.  

\paragraph{Loss weight}:

The incorporation and weighting of loss terms in PINNs have received limited attention in existing literature, despite their significant impact on performance. Many studies either merge loss terms without explicit weights~\cite{Raissi01,Jagtap2020,Wang2021} or provide little detail on their assignment~\cite{Mao2022,Chiu2022,Billah2023}, leaving a gap in clear guidelines. Recent work highlights the importance of proper weighting: Cai et al.~\cite{Mao2022} emphasize the need to balance data fitting and physics consistency, while Cuomo et al.~\cite{Cola2022} point out that multiple weighted losses complicate hyperparameter tuning, motivating the development of dedicated tools and libraries.

%%%%%%%%%%%%%%%%%%%%%%%%%%%%%%%%%%%%%%%%%%%%%%%%%%%%%%
%                   Methodology
%%%%%%%%%%%%%%%%%%%%%%%%%%%%%%%%%%%%%%%%%%%%%%%%%%%%%%

\section{Methodology}\label{method}

Physics-informed neural networks(PINNs) \cite{Raissi01} are neural networks that act as explicit functions to describe implicit governing equations of a system. Such neural networks take in independent variables of the system as the input of the neural network, and dependent variables are the outputs of the neural networks. In FIG. 4 act as explicit functions to describe some physics quantity u where the independent variables are position x y and t based on given implicit governing equations of a system and boundary conditions.
\begin{figure}[htbp!]
    \centering
    \includegraphics[width=0.6\textwidth]{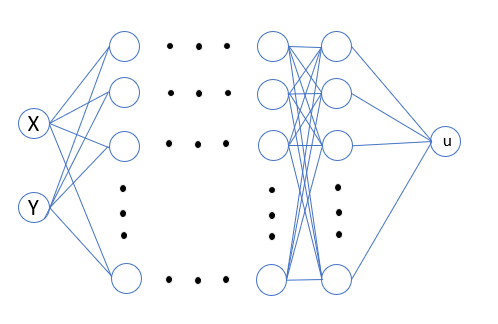}
    \caption{PINN}
\end{figure}

The loss function $\mathcal{L}$ in PINNs is defined as a weighted sum of different loss components. The differential-equation loss $\mathcal{L}_{DE}$ enforces the governing equations, the boundary-condition loss $\mathcal{L}_{BC}$ enforces boundary conditions, and the initial-condition loss $\mathcal{L}_{IC}$ enforces initial conditions. The general form is  

\begin{equation}
  \mathcal{L} = \lambda_{DE}\mathcal{L}_{DE} + \lambda_{BC}\mathcal{L}_{BC} + \lambda_{IC}\mathcal{L}_{IC}.
\end{equation}

In this thesis, PINNs are applied to three problems: two-dimensional conduction, two-dimensional convection–diffusion, and steady two-dimensional lid-driven cavity flow. For the conduction and convection–diffusion problems, the independent variables are the spatial coordinates $(x,y)$ and the dependent variable is temperature $T$. For the lid-driven cavity problem, the independent variables are also $(x,y)$, while the dependent variables are velocity components $(u,v)$ and pressure $p$.  

\subsection{Define loss function L for the PINNs: Numerical Differentiation(CDS)}\label{sec2.1}
One commonly used measure to compute this discrepancy is the mean square error (MSE) (eq(7)), which is empirically popular and employed to calculate the loss in this study.

\begin{equation}
    MSE(\hat{y},y) = \frac{1}{n}\sum_{i=1}^{n} \big(\hat{y_i}-y_i\big)^2 = \frac{1}{n}\sum e^2_i
\end{equation}

Boundary-condition loss component $\mathcal{L}_{BC}$ is defined by the residue of boundary conditions of the defined problem where 

\begin{equation}
\mathcal{L}_{BC} = \frac{1}{|\partial\Omega|} \sum_{i=1}^{|\partial\Omega|} \left( f_i - g_i \right)^2
\end{equation}
$\mathcal{i}$:index of sampling point on domain boundary $\partial\Omega$ \\
|$\partial\Omega$|:total number of sampling points on domain boundary $\partial\Omega$
\\
f: Neural network’s prediction of boundary\\
g: Boundary condition\\
Boundary-condition loss component $\mathcal{L}_{BC}$ can be either Dirichlet-boundary-condition loss component $\mathcal{L}_{DBC}$ or Neumann-boundary-condition loss component $\mathcal{L}_{NBC}$. When there both boundary condition is defined by the problem, $\mathcal{L}_{BC}$ along with its corresponding weight $\lambda_{BC}$ will be redefined as the weighted sum of loss components defined by these boundary conditions, where

\begin{equation}
    \lambda_{BC}\mathcal{L}_{BC} = \lambda_{DBC}\mathcal{L}_{DBC} + \lambda_{NBC}\mathcal{L}_{NBC}
\end{equation}

\subsection{Determine loss weight $\lambda$ for the PINNs:}\label{sec2.2}

Many studies either assign the same value to $\lambda$ or leave it undefined. Using a single $\lambda$ often leads PINNs to produce non-physical solutions, as different  $\lambda$ combinations directly affect the results. Without clear guidelines, reproducibility becomes difficult. In this section, we present strategies for setting $\lambda$, aiming to improve reproducibility and enable systematic comparisons.

\subsubsection{Dimensional Analysis:}
One of our key objectives is to analyze the order of magnitude of the loss components to balance their relative importance in the neural network. To prevent smaller-magnitude terms from being neglected, we assign larger weight parameters $\lambda$ to them and smaller weights to larger-magnitude terms. The specific values are determined through an order-of-magnitude analysis. The loss components are defined in Section~II.A, with their respective magnitudes evaluated in Sections~III.A.1--III.A.3.

\subsubsection{Investigation of Different Ratios of Loss Weight in Physics-Informed Neural Networks :}
We investigate three different weighting schemes for the loss components in Physics-Informed Neural Networks (PINNs). These schemes are derived in Sections~III.A.1--III.A.3, and solutions obtained from PINNs trained with the corresponding loss weights are compared.

The three schemes are as follows:

1. Equal weights: all loss weights are assigned the same value. This commonly used approach is denoted by the subscript ``0'', e.g $\hat{T}_0$, $\hat{u}_0$, $\hat{p}_0$.

2. Order-of-magnitude balancing: loss weights $\lambda$ are determined from the ratio of magnitudes of quantifiable terms in each loss component. For the heat-conduction PINN, the solutions are denoted with the subscript ”{$NM^2$}", e.g, $\hat{T}_{NM_2}$, $\hat{u}_{NM_2}$, $\hat{p}_{NM_2}$.

3. Relaxed order-of-magnitude balancing: a relaxation factor is introduced by taking the square root of the ratio, acknowledging that unquantifiable terms vary alongside quantifiable ones. This exploratory scheme is denoted with the subscript ``NM'', e.g., $\hat{T}_{NM}$, $\hat{u}_{NM}$, $\hat{p}_{NM}$.

By comparing different ratios of loss weights, we evaluate their impact on the performance of PINNs in capturing the underlying physics and producing accurate results. The choice of $\lambda$ is critical, as it controls the relative importance of each loss component in guiding the network’s training.

We compare different loss-weight ratios to assess their effect on PINN performance in capturing physics and producing accurate results. The choice of $\lambda$ is critical, as it governs the balance among loss components during training. In Section~III, we analyze three weighting schemes, discussing their effectiveness and limitations. Although no single optimal strategy is identified, the study offers insights into $\lambda$-scaling and its role in improving the accuracy and stability of PINNs for CFD applications.

\subsection{Method to increase model complexity for PINNs}\label{sec2.3}
Model complexity is critical to the accuracy of Physics-Informed Neural Networks (PINNs). As sampling points and problem difficulty increase, more trainable parameters are required to achieve physics-consistent solutions. In this thesis, a five-layer network with 64 neurons per layer is adopted as the default, balancing efficiency and capacity.

PINN complexity is primarily determined by network architecture. Increasing neurons is often more effective than adding layers, consistent with Fourier’s principle that any function can be represented as a series of sines \cite{Zygmund2002}. In Section~III.B.1, we compare different configurations to highlight the trade-offs between complexity, accuracy, and computational cost, providing guidance for selecting architectures in CFD applications.

\subsection{Benchmark}\label{sec2.4}

This study apply finite difference method (FDM) to obtain numerical result for benchmark. Discretization of differential equations is based on Taylor expansion. Derivatives of a function, e.g, $\frac{\partial{u}(x, y, t)}{\partial{t}}$, on discretized domain utilize Taylor expansion of functions adjacent points in the direction independent variable, e.g, x, y, t. Function of an adjacent points are described as:
\begin{equation}
\begin{aligned}
f'(x) &= \frac{f(x+h) - f(x-h)}{2h}
\end{aligned}
\end{equation}
and we obtain the second derivative 
\begin{equation}
\begin{aligned}
f''(x) &= \frac{f(x+h) + f(x-h) - 2\times f(x)}{h^2}
\end{aligned}
\end{equation}

\subsubsection{Conduction:}
Conduction problem is depicted via eq(27). The discretized form
of governing equation is
\begin{equation}
    T_{i,j} = \frac{T_{i+1,j}+T_{i,j+1}+T_{i-1,j}+T_{i,j-1}}{4}
\end{equation}

The iterative process run on problem domain $\Omega$ on collocated grid.  Gauss Seidel iteration is applied, and the convergence criterion is when

\begin{equation}
\max_{(i,j)\in\Omega} 
\frac{T^{n+1}_{i,j} - T^{n}_{i,j}}{10^{-20} + T^{n}_{i,j}}
< 10^{-6}
\end{equation}

\subsubsection{convection-and-diffusion:}
Convection-and-diffusion problem is depicted via eq(39). The discretized form of governing equation is
\begin{equation}
T_{i,j} = 
\frac{
T_{i+1,j} + T_{i-1,j} + T_{i,j+1} + T_{i,j-1}
- \tfrac{1}{2} Pe \cdot \left[ (T_{i+1,j} - T_{i-1,j}) + (T_{i,j+1} - T_{i,j-1}) \right]
}{4}
\end{equation}

The iterative process is performed on the problem domain $\Omega$ using a collocated grid. The Gauss-Seidel iteration is applied, and convergence is achieved when

\begin{equation}
\max_{(i,j)\in\Omega} 
\frac{T^{n+1}_{i,j} - T^{n}_{i,j}}{10^{-20} + T^{n}_{i,j}}
< 10^{-6}
\end{equation}

\subsubsection{Lid-driven-cavity:}
Flow characteristics of viscous incompressible flow is depicted via the law of conservation of momentum, for fluid it is the Navier-Stokes equation eq(51), eq(52), and the law of conservation of mass, or the continuity(eq(53). 

Governing equations can be written in vector form, where the momentum equations in vector form is described as

\begin{equation}
\mathbf{v}_t + \mathbf{v} \left( \nabla \cdot \mathbf{v} \right)
= - \nabla p + \nu \nabla^2 \mathbf{v}
\end{equation}
Continuity in vector form is described as

\begin{equation}
\nabla \cdot \mathbf{v} = 0
\end{equation}
The flow domain is a square region and a uniform grid is applied, and compute for u, v and p on every grid point ${c_{ij}}$. To avoid checkerboard distribution in the process of computation, staggered grid is applied, and the grid is then collocated onto the uniform grid on the flow domain.

This study apply finite difference method (FDM) for discretization.	The discretized form of momentum equation in x-direction is

\begin{equation}
\begin{split}
\frac{u_{i,j}^{n+1}-u_{i,j}^{n}}{\Delta t}
&\;+\; u_{i,j}\,\frac{u_{i+1,j}-u_{i-1,j}}{2\Delta x}
     + \hat{v}_{i,j}\,\frac{u_{i,j+1}-u_{i,j-1}}{2\Delta y} \\
&= \frac{p_{i+1,j}-p_{i,j}}{\Delta x}
 \;+\; \frac{1}{Re}\!\left(
      \frac{u_{i-1,j}+u_{i+1,j}-2u_{i,j}}{\Delta x^{2}}
    + \frac{u_{i,j-1}+u_{i,j+1}-2u_{i,j}}{\Delta y^{2}}
    \right)
\end{split}
\end{equation}

$\hat{v}_{i,j}$ denotes $\mathcal{v}$ on the grid point $\mathcal{u}_{i,j}$, a grid point on the grid for $\mathcal{u}$ where,

\begin{equation}
\hat{v}_{i,j} =
\frac{v_{i,j} + v_{i+1,j} + v_{i,j-1} + v_{i+1,j-1}}{4}
\end{equation}
and the discretized form of momentum equation in y-direction is
\begin{align}
\frac{v_{i,j}^{n+1}-v_{i,j}^{n}}{\Delta t}
&\;+\; \hat{u}_{i,j}\,\frac{v_{i+1,j}-v_{i-1,j}}{2\Delta x}
   + v_{i,j}\,\frac{v_{i,j+1}-v_{i,j-1}}{2\Delta y} \notag \\[6pt]
&= \frac{p_{i,j+1}-p_{i,j}}{\Delta y}
 \;+\; \frac{1}{Re}\!\left(
      \frac{v_{i-1,j}+v_{i+1,j}-2v_{i,j}}{\Delta x^{2}}
    + \frac{v_{i,j-1}+v_{i,j+1}-2v_{i,j}}{\Delta y^{2}}
    \right) 
\end{align}

$\hat{u}_{i,j}$ denotes u on the grid point $\mathcal{v}_{i,j}$, where

\begin{equation}
\hat{u}_{i,j} = \frac{u_{i-1,j+1} + u_{i,j+1} + u_{i-1,j} + u_{i,j}}{4}
\end{equation}

Algorithm adopted in this work is projection method. Projection method split eq(16) into
\begin{equation}
\frac{\mathbf{v}^* - \mathbf{v}^n}{\Delta t}
+ \mathbf{C}(\mathbf{v}^n)
= -\nabla p^n + \mathbf{D}(\mathbf{v}^n)
\end{equation}
\begin{equation}
\frac{\mathbf{v}^{**} - \mathbf{v}^*}{\Delta t}
= \nabla p^n
\end{equation}
\begin{equation}
\frac{\mathbf{v}^{n+1} - \mathbf{v}^{**}}{\Delta t}
= -\nabla p^{n+1}
\end{equation}
, the sum of the LHP and the RHP of 3 equations add up to be the momentum equation(17). Take divergence of eq(24), we have

\begin{equation}
\nabla \cdot \frac{\mathbf{v}^{n+1} - \mathbf{v}^{**}}{\Delta t}
= -\nabla^2 p^{n+1}
\end{equation}

As a part of the algorithm, velocity field under each time-step reach divergence free under every time step, i.e, $\nabla \cdot \mathbf{v}^{n+1}$ = 0, therefore, we have 
\begin{equation}
\nabla \cdot \frac{\mathbf{v}^{**}}{\Delta t} = \nabla^2 p^{n+1}
\end{equation}

Projection method utilize these fraction equations of eq(17) to solve for p, u and v at each time-step. For each time-step, the procedure may be described into 4 steps: first, solve $\mathbf{v}^*$ with eq(22); second, solve $\mathbf{v}^{**}$ with eq(23); third, iterative update $p^{n+1}$ with eq(26); and last, solve $v^{n+1}$ with eq(24).

\begin{algorithm}[H]
\caption{Projection method}
\KwResult{Velocity field and pressure field under each time-step}
initialization\;
\While{not yet reach steady state}{
    solve $\mathbf{v}^*$ with eq(22)\;
    solve $\mathbf{v}^{**}$ with eq(23)\;
    take initial guess of $p^{n+1}$\;
    \While{Poisson equation not solve}{
        iteratively update $p^{n+1}$ with eq(26)\;
    }
    solve $\mathbf{v}^{n+1}$ with eq(24)\;
    collocate $\mathbf{v}^{n+1}$\;
}
\end{algorithm}

\subsection{Framework for deep-learning: Pytorch}\label{sec2.5}
This study employs PyTorch, an open-source machine learning framework for Python that supports efficient research prototyping and deployment. Developed by Facebook AI Research, PyTorch builds on the Torch library with a Python interface while retaining the optimized C backend and GPU acceleration, ensuring both flexibility and performance.

\subsection{Integration of numerical solver and learning based method}\label{sec2.6}
PINNs are newly introduced in this study and not yet integrated with existing CFD solvers. However, such integration is essential for practical applications, as numerical solvers are typically developed in C/C++ or FORTRAN, whereas learning-based methods are commonly implemented in Python. Two coupling strategies exist: embedding Python-based learning methods into C++ solvers, or incorporating C++ solvers into Python frameworks. In Section~III.C, we compare these approaches and explain why integrating Python-based methods into C++ solvers is the preferred choice..

%%%%%%%%%%%%%%%%%%%%%%%%%%%%%%%%%%%%%%%%%%%%%%%%%%%%%%
%                   Experimental study
%%%%%%%%%%%%%%%%%%%%%%%%%%%%%%%%%%%%%%%%%%%%%%%%%%%%%%

\section{Experimental study}\label{method}
\subsection{Solution obtain by PINN with varying loss weight at h = $\frac{1}{10}$, $\frac{1}{30}$ and $\frac{1}{50}$}\label{sec3.1}
In this section, we compare the solutions obtained from PINNs trained with varying loss weight to determine physic quantity on equidistant-spaced grid with spacing h = $\frac{1}{10}$, $\frac{1}{30}$ and $\frac{1}{50}$. The PINN architecture, training configuration, and also the training cost are summarized in Table I.

\begin{table}[h]
\centering
\renewcommand{\arraystretch}{1.2}
\setlength{\tabcolsep}{6pt} % 調整欄間距
\begin{tabular}{l c c c}
\hline
\textbf{Problem} &
\makecell{Conduction\\(Section III.A.1)} &
\makecell{Convection and Diffusion\\(Section III.A.2)} &
\makecell{Lid-driven cavity\\(Section III.A.3)} \\
\hline
Governing equations & eq(27) & eq(39) & eq(51), eq(52), eq(53) \\
Loss functions      & eq(29) & eq(43) & eq(57) \\
\hline
PINN architecture &
$(x,y)\text{--}64\text{--}64\text{--}64\text{--}64\text{--}(\hat{T})$ &
$(x,y)\text{--}64\text{--}64\text{--}64\text{--}64\text{--}(\hat{T})$ &
$(x,y)\text{--}64\text{--}20\text{--}20\text{--}20\text{--}[\hat{u},\hat{v},\hat{p}]$ \\
Optimizer & Adam \cite{Kingma2014} & Adam & Adam \\
$h$ & $\tfrac{1}{10},\ \tfrac{1}{30},\ \tfrac{1}{50}$
    & $\tfrac{1}{10},\ \tfrac{1}{30},\ \tfrac{1}{50}$
    & $\tfrac{1}{10},\ \tfrac{1}{30},\ \tfrac{1}{50}$ \\
Training sample & 121 / 961 / 2601 & 121 / 961 / 2601 & 121 / 961 / 2601 \\
Max training iteration & 50{,}000 & 50{,}000 & 80{,}000 \\
Initial learning rate   & $10^{-3}$ & $10^{-3}$ & $10^{-2}$ \\
Learning rate decay     & 0.8 / 1000 epoch & 0.8 / 1000 epoch & 0.8 / 1000 epoch \\
Total training cost     & 4.1 min & 4.1 min & 12.3 min \\
\hline
\end{tabular}

\caption{PINN setup. The numbers denote hidden-layer widths. 
For example, $(x,y)\text{--}64\text{--}64\text{--}64\text{--}64\text{--}(\hat{T})$ indicates two inputs $(x,y)$, four hidden layers with 64 nodes each, and a single output $\hat{T}$. 
Hidden layers use ``sine'' activation; the output layer uses ``linear''.}
\label{table:pinn-setup}
\end{table}

\subsubsection{Condution:}
In this section, we compare the solutions obtained from PINNs trained with varying loss weight. A two-dimensional conduction problem is governed by the following differential equation, 
\begin{equation}
\frac{\partial^2 T}{\partial x^2} + \frac{\partial^2 T}{\partial y^2} = 0
\end{equation}
where the domain boundary $\partial\Omega$ is defined with boundary condition g where
\begin{equation}
g(x,y) =
\begin{cases}
0, & \text{if } x=0,1;\; y=0 \\
1, & \text{if } y=1
\end{cases}
\end{equation}
\begin{figure}[htbp!]
    \centering
    \includegraphics[width=0.6\textwidth]{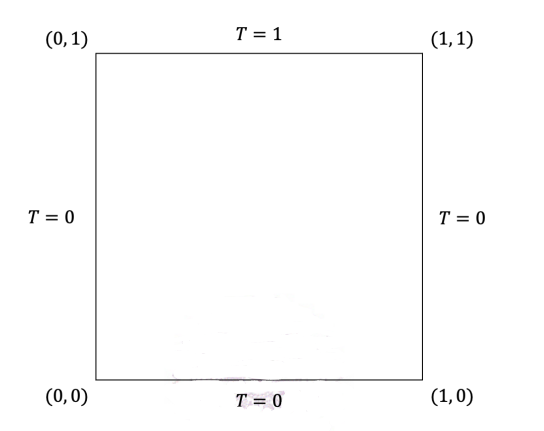}
    \caption{Problem domain and boundary conditions of conduction problem}
\end{figure}

\paragraph{Define loss function $\mathcal{L}$}: 

A two-dimensional conduction problem is governed by eq(27). The conduction PINN is trained with loss function $\mathcal{L}$ composed of differential loss component $\mathcal{L}_{DE}$ and Dirichlet boundary condition loss component $\mathcal{L}_{DBC}$, each loss component is multiplied with a corresponding loss weight $\lambda$, which is defined as
 \begin{equation}
     \mathcal{L}=\lambda_{DE}\mathcal{L}_{DE}+\lambda_{DBC}\mathcal{L}_{DBC}
 \end{equation}
where loss components of conduction are defined by numerical differentiation(ND) with central differencing scheme(CDS), when compute with MSE, differential-equation loss component $\mathcal{L}_{DE}$ is defined as
\begin{equation}
\mathcal{L}_{DE} = \frac{\left\lVert \nabla^2 T \right\rVert^{2}_{\Omega}}{|\Omega|}
\end{equation}
where
\begin{equation}
\left\lVert \nabla^2 T \right\rVert^{2}_{\Omega}
= \sum_{i=1}^{N-2} \sum_{j=1}^{N-2} 
\left( \frac{T_{i+1,j} + T_{i-1,j} + T_{i,j+1} + T_{i,j-1} - 4T_{i,j}}{h^2} \right)^{2}
\end{equation}
and temperature boundary condition eq(28) is written into the Dirichlet boundary-condition loss component $\mathcal{L}_{DBC}$ is defined as
\begin{equation}
\mathcal{L}_{DBC} = \frac{\left\lVert T - g \right\rVert^{2}_{\partial \Omega}}{|\partial \Omega|}
\end{equation}

\paragraph{Dimensional Analysis}:

Two-dimensional conduction problem is governed by eq(27). The conduction PINN is trained with loss function composed of differential loss component $\mathcal{L}_{DE}$ and Dirichlet boundary condition loss component $\mathcal{L}_{DBC}$, each loss component is multiplied with a corresponding loss weight $\lambda$ (eq(29)). To determine the proper ratio between each loss function, analyze on order of magnitude of loss component is considered. Order of magnitude of differential equation loss component is
\begin{equation}
\mathcal{L}_{DE} = \frac{\left\lVert \nabla^2 T \right\rVert^{2}_{\Omega}}{|\Omega|}
    \;\sim\; \left( \frac{T_{i+1,j} + T_{i-1,j} + T_{i,j+1} + T_{i,j-1} - 4T_{i,j}}{h^2} \right)^{2}
    \;\sim\; \frac{[T]^2}{[h]^4}
\end{equation}
Order of magnitude of Dirichlet boundary condition loss component is
\begin{equation}
\mathcal{L}_{DBC} = \frac{\left\lVert T-g \right\rVert^{2}_{\partial \Omega}}{|\partial \Omega|}
    \;\sim\; [T]^2
\end{equation}

\paragraph{Investigation of Different Ratios of Loss Weight for Physics-Informed Neural Networks}:
Loss weights $\lambda_s$ in eq(29) control the contribution of each different component, we investigate three different weighting schemes for the loss components in Physics-Informed Neural Networks (PINNs). The three schemes are: when the loss weights are given the same value, when the ratio of loss weights is determined from the analysis of the order of magnitude, and when the square root of the ratio is employed as a relaxation factor. The first scheme represents the most commonly used approach for setting loss weights in PINNs, where

\begin{equation}
    \lambda_{DE} : \lambda_{DBC} = 1:1
\end{equation}
Solutions obtained from PINNs trained ratio set with the first scheme will be denoted with a subscript ”0”, i.e, $\hat{T}_0$

The second scheme aims to balance the order of magnitude of the loss components by determining the loss weights based on the magnitude of quantifiable terms(eq(33), eq(34)) within the loss components, where
\begin{align}
[\lambda_{DE}\mathcal{L}_{DE}] &\sim [\lambda_{DBC}\mathcal{L}_{DBC}] \\[6pt]
[\lambda_{DE}] : [\lambda_{DBC}] &\approx [h]^4 : 1
\end{align}
Solutions obtained from PINNs trained ratio set with the second scheme will be denoted with a subscript ”$NM^2$”, i.e, $\hat{T}_{NM^2}$.

The third scheme introduces a relaxed version of the second scheme, taking into consideration that the magnitude of unquantifiable terms tends to change alongside the quantifiable terms. As a result, some relaxation is applied to the determined ratio, with the square root being used in this research as a form of relaxation, where

\begin{equation}
    \lambda_{DE}:\lambda_{DBC} = h^2:1
\end{equation}
Solutions obtained from PINNs trained ratio set with the third scheme will be denoted with a subscript ”NM ”, i.e, $\hat{T}_{NM}$.

\paragraph{Solutions obtained from PINNs}:

FIG. 5 shows solutions obtained from PINNs and benchmark solution obtained from finite difference method $T_{FDM}$ at y =0.5. FIG. 5(a) shows that $\hat{T}_{0}$, $\hat{T}_{NM}$ and $\hat{T}_{NM^2}$ agree with benchmark solution when h = $\frac{1}{10}$ FIG. 5(b) shows that $\hat{T}_{0}$, $\hat{T}_{NM}$ and $\hat{T}_{NM^2}$ agree with benchmark solution when h = $\frac{1}{30}$; FIG. 5(c) show that only $\hat{T}_{NM}$ agree with benchmark solution when h = $\frac{1}{50}$.

Corresponding mean square error(MSE) can be found in Table II. The efficacy and accuracy of PINN trained with the third weighting scheme(eq(38)) is thus demonstrated in this test problem.
\begin{figure}[htbp!]
    \centering
    % 第一張子圖
    \begin{subfigure}{0.32\textwidth}
        \centering
        \includegraphics[width=\textwidth]{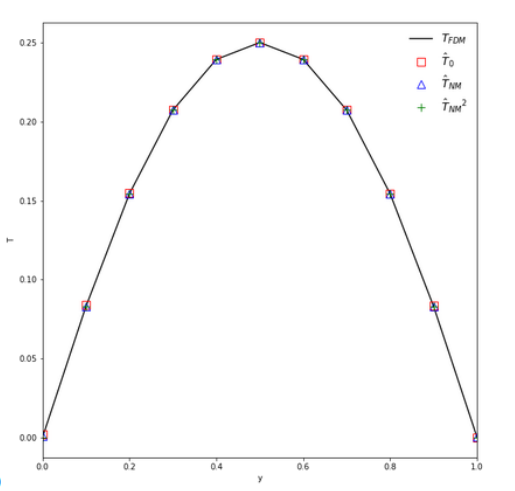}
        \caption{}
    \end{subfigure}
    % 第二張子圖
    \begin{subfigure}{0.32\textwidth}
        \centering
        \includegraphics[width=\textwidth]{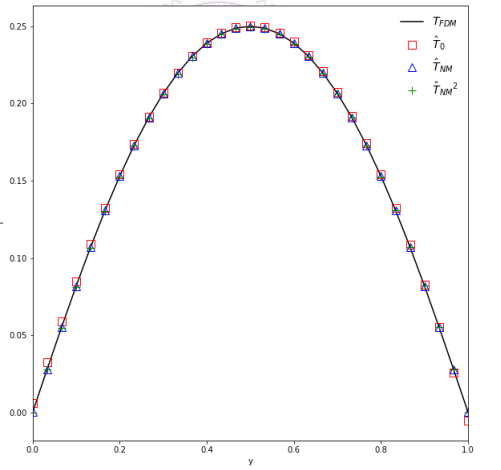}
        \caption{}
    \end{subfigure}
    % 第三張子圖
    \begin{subfigure}{0.32\textwidth}
        \centering
        \includegraphics[width=\textwidth]{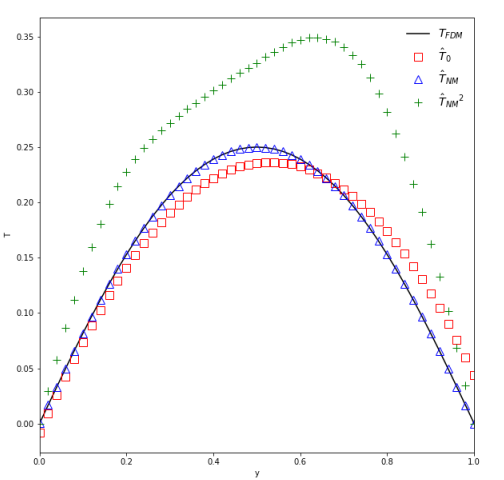}
        \caption{}
    \end{subfigure}

    \caption{PINN’s prediction of temperature distribution at y = 0.5 at (a) $\frac{1}{10}$ (b) $\frac{1}{30}$ (c) $\frac{1}{50}$ }
\end{figure}

\begin{table}[htbp!]
\centering
\renewcommand{\arraystretch}{1.2} % 行距微調
\begin{tabular}{c|c|c|c}
\hline
($\times 10^{-3}$) & $\hat{T}_0$ & $\hat{T}_{NM}$ & $\hat{T}_{NM^2}$ \\
Loss weight & eq(35) & eq(38) & eq(37) \\
\hline
$h = \tfrac{1}{10}$ & 4.143 & 4.132 & 4.132 \\
$h = \tfrac{1}{30}$ & 8.786 & 5.203 & 5.204 \\
$h = \tfrac{1}{50}$ & {\color{red} 3.755} & 0.192 & {\color{red} 3.437} \\
\hline
\end{tabular}
\caption{Mean square error (MSE) of solutions obtained from conduction PINN.}
\label{tab:conduction-MSE}
\end{table}

The following figures show results of conduction PINN when h = $\frac{1}{10}$ FIG. 6 shows distribution of $\hat{T}_{0}$, $\hat{T}_{NM}$ and $\hat{T}_{NM^2}$ in the  problem domain and their error from $T_{FDM}$.
\begin{figure}[htbp!]
    \centering
    \includegraphics[width=0.9\textwidth]{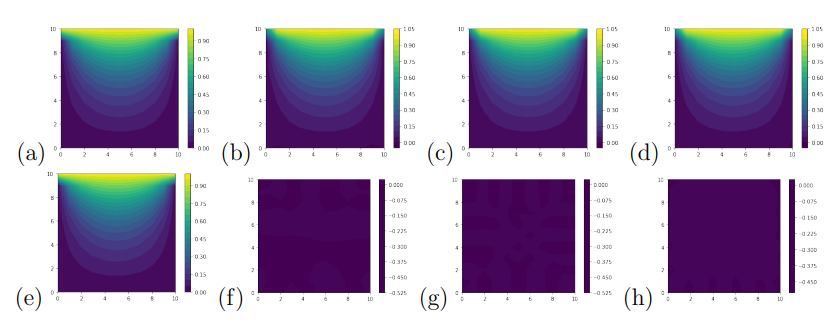}
    \caption{(a) FDM solution $T_{FDM}$ 
(b) PINN’s prediction of temperature with $\lambda_{DE}:\lambda_{DBC}=1:1 \; \hat{T}_0$ 
(c) PINN’s prediction of temperature with $\lambda_{DE}:\lambda_{DBC}=h^2:1 \; \hat{T}_{NM}$ 
(d) PINN’s prediction of temperature with $\lambda_{DE}:\lambda_{DBC}=h^4:1 \; \hat{T}_{NM^2}$ 
(e) $T_{FDM}$ 
(f) Error between $\hat{T}_0$ and $T_{FDM}$ 
(g) Error between $\hat{T}_{NM}$ and $T_{FDM}$ 
(h) Error between $\hat{T}_{NM^2}$ and $T_{FDM}$.}
\end{figure}

The following figures show results of conduction PINN when h = $\frac{1}{30}$ FIG. 7 shows distribution of $\hat{T}_{0}$, $\hat{T}_{NM}$ and $\hat{T}_{NM^2}$ in the  problem domain and their error from $T_{FDM}$.
\begin{figure}[htbp!]
    \centering
    \includegraphics[width=0.9\textwidth]{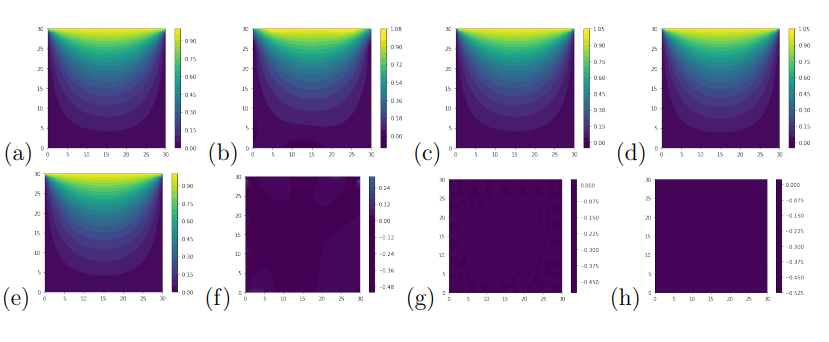}
    \caption{Temperature distribution at $h=\tfrac{1}{30}$ with (a) FDM solution $T_{FDM}$ (b) PINN’s prediction of temperature with $\lambda_{DE}:\lambda_{DBC}=1:1$ $\hat{T}_0$ (c) PINN’s prediction of temperature with $\lambda_{DE}:\lambda_{DBC}=h^2:1$ $\hat{T}_{NM}$ (d) PINN’s prediction of temperature with $\lambda_{DE}:\lambda_{DBC}=h^4:1$ $\hat{T}_{NM^2}$ (e) $T_{FDM}$ (f) Error between $\hat{T}_0$ and $T_{FDM}$ (g) Error between $\hat{T}_{NM}$ and $T_{FDM}$ (h) Error between $\hat{T}_{NM^2}$ and $T_{FDM}$.}
\end{figure}

The following figures show results of conduction PINN when h = $\frac{1}{50}$ FIG. 8 shows distribution of $\hat{T}_{0}$, $\hat{T}_{NM}$ and $\hat{T}_{NM^2}$ in the  problem domain and their error from $T_{FDM}$.
\begin{figure}[htbp!]
    \centering
    \includegraphics[width=0.9\textwidth]{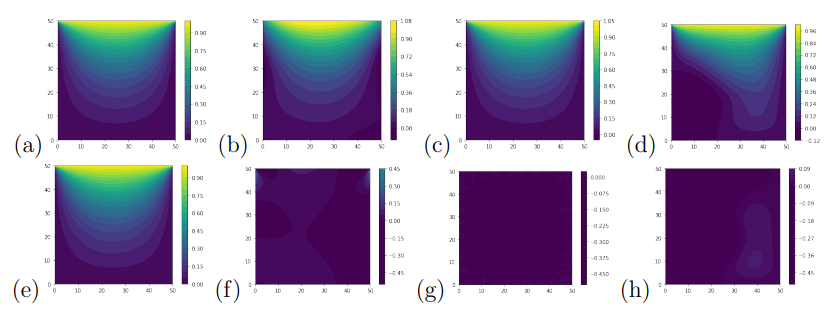}
    \caption{Temperature distribution at $h=\tfrac{1}{50}$ with (a) FDM solution $T_{FDM}$ (b) PINN’s prediction of temperature with $\lambda_{DE}:\lambda_{DBC}=1:1$ $\hat{T}_0$ (c) PINN’s prediction of temperature with $\lambda_{DE}:\lambda_{DBC}=h^2:1$ $\hat{T}_{NM}$ (d) PINN’s prediction of temperature with $\lambda_{DE}:\lambda_{DBC}=h^4:1$ $\hat{T}_{NM^2}$ (e) $T_{FDM}$ (f) Error between $\hat{T}_0$ and $T_{FDM}$ (g) Error between $\hat{T}_{NM}$ and $T_{FDM}$ (h) Error between $\hat{T}_{NM^2}$ and $T_{FDM}$.}
\end{figure}

\subsubsection{Convection and diffusion}
In this section, we compare the solutions obtained from PINNs trained with varying loss weight in Convection-and-diffusion PINN when Pe = 10 and 100 respectively. A two dimensional convection-and-diffusion problem is governed by the following differential equation, 

\begin{equation}
u \frac{\partial T}{\partial x} + v \frac{\partial T}{\partial y}
= \Gamma \left( \frac{\partial^2 T}{\partial x^2} + \frac{\partial^2 T}{\partial y^2} \right)
\end{equation}
suppose
\begin{equation}
\Gamma = \frac{\rho u h}{Pe}, \quad u = v = 1, \; \rho = 1
\end{equation}
eq(39) is rewritten into
\begin{equation}
\frac{Pe}{h} \left( \frac{\partial T}{\partial x} + \frac{\partial T}{\partial y} \right)
= \frac{\partial^2 T}{\partial x^2} + \frac{\partial^2 T}{\partial y^2}
\end{equation}
, domain boundary $\partial \Omega$ is defined with boundary condition g where
\begin{equation}
g(x,y) =
\begin{cases}
0, & \text{if } x=0,1;\; y=0 \\
1, & \text{if } x=1;\; y=1
\end{cases}
\end{equation}
\begin{figure}[htbp!]
    \centering
    \includegraphics[width=0.6\textwidth]{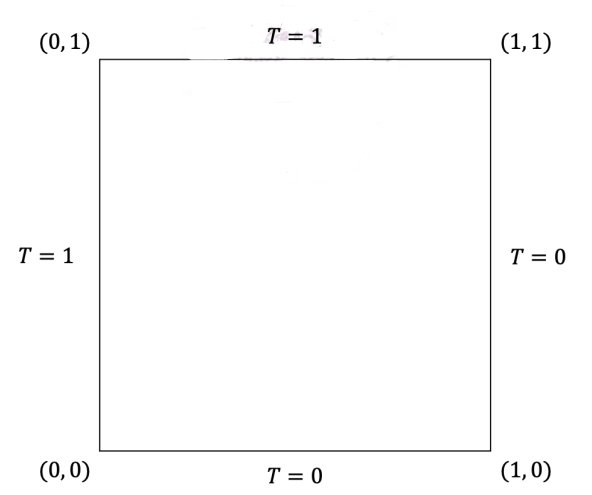}
    \caption{ Problem domain and boundary conditions of convection-and-diffusion problem}
\end{figure}

\paragraph{Define loss function $\mathcal{L}$}:

A two-dimensional convection-and-diffusion problem is governed by eq(39). The convection-and-diffusion PINN is trained with loss function $\mathcal{L}$ composed of differential loss composed of differential loss component $\mathcal{L}_{DE}$ and Dirichlet boundary condition loss component $\mathcal{L}_{DBC}$ each loss component is multiplied with a corresponding loss weight $\lambda$, which is defined as

\begin{equation}
    \mathcal{L} = \lambda_{DE}\mathcal{L}_{DE}+\lambda_{DBC}\mathcal{L}_{DBC}
\end{equation}
where loss components of conduction are defined by numerical differentiation(ND) with central differencing scheme(CDS), when compute with MSE, differential equation loss component $\mathcal{L}_{DE}$ is defined as

\begin{equation}
\begin{split}
\left\lVert \nabla^2 T - \frac{Pe}{h} \nabla \cdot T \right\rVert^2_{\Omega} 
&= \sum_{i=1}^{N-2} \sum_{j=1}^{N-2} 
\left( \frac{T_{i+1,j} + T_{i-1,j} + T_{i,j+1} + T_{i,j-1} - 4T_{i,j}}{h^2} \right. \\
&\quad \left. - \frac{Pe}{h} \cdot \frac{(T_{i+1,j}-T_{i-1,j}) + (T_{i,j+1}-T_{i,j-1})}{2h} \right)^2
\end{split}
\label{eq:3.19}
\end{equation}

and temperature boundary condition eq(42) is written into Dirichlet boundary condition loss component $\mathcal{L}_{DBC}$ is defined as 

\begin{equation}
\begin{split}
\lVert T - g \rVert^2_{\partial \Omega} =
\sum_{i=0}^{N-2} \left(T_{i,0} - g_{i,0}\right)^2
+ \sum_{i=0}^{N-2} \left(T_{i,N-1} - g_{i,N-1}\right)^2 \\
+ \sum_{j=0}^{N-2} \left(T_{0,j} - g_{0,j}\right)^2
+ \sum_{j=0}^{N-2} \left(T_{N-1,j} - g_{N-1,j}\right)^2
\end{split}
\end{equation}

\paragraph{Dimensional Analysis}:

Two-dimensional convection-and-diffusion problem is governed by eq(39). The convection-and-diffusion PINN is trained with loss function composed of differential loss component $\mathcal{L}_{DE}$ and Dirichlet boundary condition loss component $\mathcal{L}_{DBC}$, each loss component is multiplied with a corresponding loss weight $\lambda$ (eq(43)). To determine the proper ratio between each loss function, analyze on order of magnitude of loss component is considered. Order of magnitude of differential equation loss component is
\begin{equation}
\mathcal{L}_{DE} = \left\| \nabla^2 T - \frac{Pe}{h}\nabla \cdot T \right\|^2_{\Omega}
  \;\sim\; \left[ \frac{Pe}{h} \cdot \frac{T_{i+1,j} - T_{i-1,j} + T_{i,j+1} - T_{i,j-1}}{2h} \right]^2
  \;\sim\; \frac{[Pe]^2 [T]^2}{[h]^4}
\end{equation}
, and order of magnitude of Dirichlet boundary condition loss component $\mathcal{L}_{DBC}$ is 
\begin{equation}
\mathcal{L}_{DBC} = \frac{\| T - g \|^2_{\partial \Omega}}{|\partial \Omega|}
  \;\sim\; [T]^2
\end{equation}

\paragraph{Investigation of Different Ratios of Loss Weight for Physics-Informed Neural Networks}:

Loss weights $\lambda_s$ in in eq(43) control the contribution of each different component, we investigate three different weighting schemes for the loss components in Physics-Informed Neural Networks (PINNs). The three schemes are: when the loss weights are given the same value, when the ratio of loss weights is determined from the analysis of the order of magnitude, and when the square root of the ratio is employed as a relaxation factor. The first scheme represents the most commonly used approach for setting loss weights in PINNs, where
\begin{equation}
    \lambda_{DE}:\lambda_{DBC} = 1:1
\end{equation}
Solutions obtained from PINNs trained ratio set with the first scheme will be denoted with a subscript ”0”, i.e, $\hat{T}_0$.

The second scheme aims to balance the order of magnitude of the loss components by determining the loss weights based on the magnitude of quantifiable terms(eq(46), eq(47)) within the loss components, where

\begin{equation}
\begin{aligned}
[\lambda_{DE}\mathcal{L}_{DE}] &\;\sim\; [\lambda_{DBC}\mathcal{L}_{DBC}] \\[6pt]
\lambda_{DE} : \lambda_{DBC} &\;\approx\; \frac{[h]^4}{[Pe]^2} : 1
\end{aligned}
\end{equation}

Solutions obtained from PINNs trained ratio set with the second scheme will be denoted with a subscript "$NM^2$", i.e, $\hat{T}_{NM^2}$.

The third scheme introduces a relaxed version of the second scheme, taking into consideration that the magnitude of unquantifiable terms tends to change alongside the quantifiable terms. As a result, some relaxation is applied to the determined ratio, with the square root being used in this research as a form of relaxation, where
\begin{equation}
\lambda_{DE} : \lambda_{DBC} = Pe^{-1}h^{2} : 1
\end{equation}

Solutions obtained from PINNs trained ratio set with the third scheme will be denoted with a subscript ”NM”, i.e, $\hat{T}_{NM}$.

\paragraph{Solution obtain by PINNs at Pe = 10}:

FIG. 10 shows solutions obtained from PINNs and benchmark solution obtained from finite difference method $T_{FDM}$ at x = 0.5 when Pe = 10. FIG. 10(a) shows that $\hat{T}_{NM}$ and $\hat{T}_{NM^2}$ agree with benchmark solution when h = $\frac{1}{10}$; FIG. 10(b) shows that $\hat{T}_{NM}$ and $\hat{T}_{NM^2}$ agree with benchmark solution when h = $\frac{1}{30}$; and FIG. 10(c) shows that $\hat{T}_{NM}$ and $\hat{T}_{NM^2}$ agree with benchmark solution when h = $\frac{1}{50}$.

Corresponding mean square error(MSE) can be found in Table III. The efficacy and accuracy of PINN trained with eq(49), eq(50) are thus demonstrated in this test problem. 
\begin{figure}[htbp!]
    \centering
    % 第一張子圖
    \begin{subfigure}{0.32\textwidth}
        \centering
        \includegraphics[width=\textwidth]{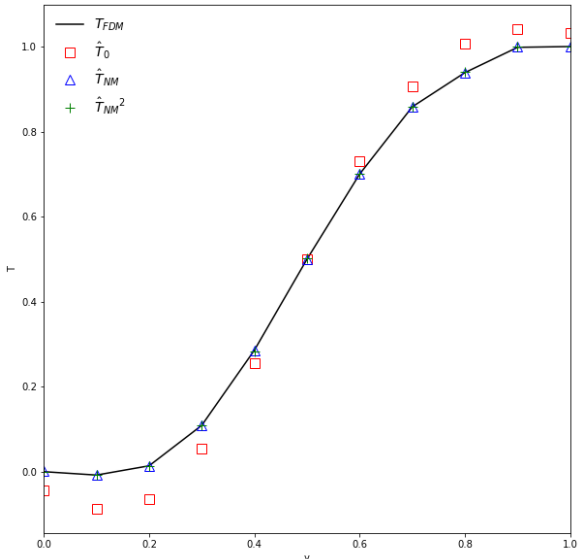}
        \caption{}
    \end{subfigure}
    % 第二張子圖
    \begin{subfigure}{0.32\textwidth}
        \centering
        \includegraphics[width=\textwidth]{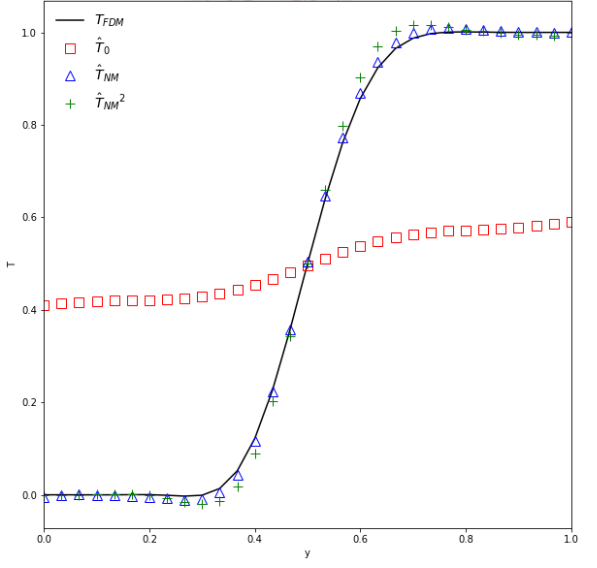}
        \caption{}
    \end{subfigure}
    % 第三張子圖
    \begin{subfigure}{0.32\textwidth}
        \centering
        \includegraphics[width=\textwidth]{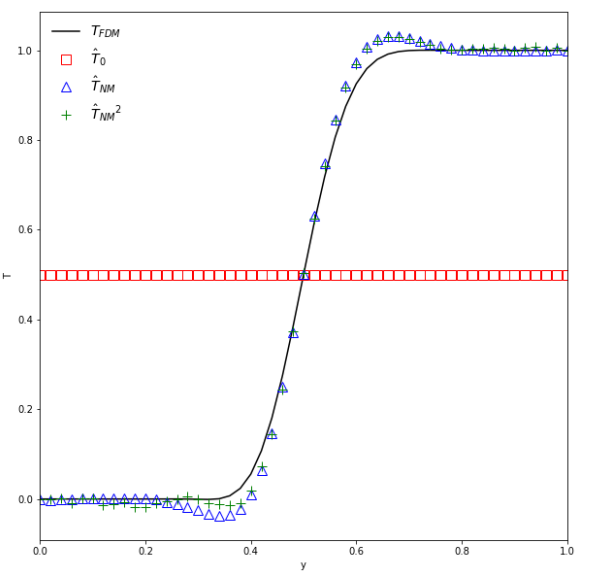}
        \caption{}
    \end{subfigure}

    \caption{ PINN’s prediction of temperature distribution at x = 0.5 at (a) $\frac{1}{10}$ (b) $\frac{1}{30}$ (c) $\frac{1}{50}$ }
\end{figure}

\begin{table}[h]
\centering
\renewcommand{\arraystretch}{1.2} % 行距微調
\begin{tabular}{c|c|c|c}
\hline
($\times 10^{-3}$) & $\hat{T}_0$ & $\hat{T}_{NM}$ & $\hat{T}_{NM^2}$ \\
Loss weight & eq(48) & eq(50) & eq(49) \\
\hline
$h = \tfrac{1}{10}$ & 8.799 & 4.133 & 4.134 \\
$h = \tfrac{1}{30}$ & {\color{red} 146.7} & 0.586 & 1.143 \\
$h = \tfrac{1}{50}$ & {\color{red} 218.2} & 1.082 & 1.155 \\
\hline
\end{tabular}
\caption{Mean square error (MSE) of solutions obtained from \newline convection-and-diffusion PINN at $Pe = 10$.}
\label{tab:conv-diff-MSE-Pe10}
\end{table}

The following figures show results of PINN with h = $\frac{1}{10}$ FIG. 11 shows distribution of $\hat{T}_0$, $\hat{T}_{NM}$ and $\hat{T}_{NM^2}$ in the problem domain and their error from $T_{FDM}$.
\begin{figure}[htbp!]
    \centering
    \includegraphics[width=0.9\textwidth]{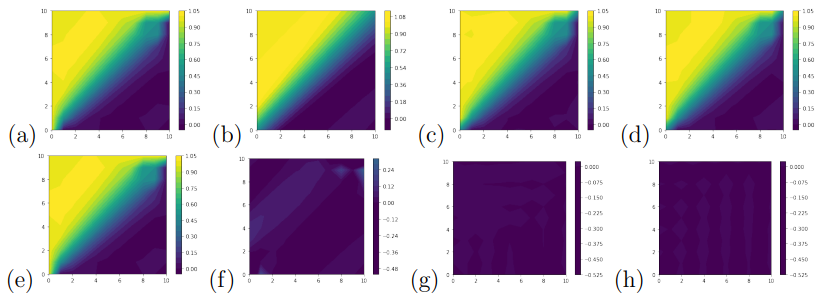}
    \caption{Temperature distribution at $Pe = 10, \; h = \tfrac{1}{10}$ with (a) FDM solution $T_{FDM}$ (b) PINN’s prediction of temperature with $\lambda_{DE} : \lambda_{DBC} = 1 : 1 \; \hat{T}_0$ (c) PINN’s prediction of temperature with $\lambda_{DE} : \lambda_{DBC} = Pe^{-1}h^2 : 1 \; \hat{T}_{NM}$ (d) PINN’s prediction of temperature with $\lambda_{DE} : \lambda_{DBC} = Pe^{-2}h^4 : 1 \; \hat{T}_{NM^2}$ (e) $T_{FDM}$ (f) Error between $\hat{T}_0$ and $T_{FDM}$ (g) Error between $\hat{T}_{NM}$ and $T_{FDM}$ (h) Error between $\hat{T}_{NM^2}$ and $T_{FDM}$.}
\end{figure}

The following figures show results of PINN with h = $\frac{1}{10}$ FIG.12 shows distribution of $\hat{T}_0$, $\hat{T}_{NM}$ and $\hat{T}_{NM^2}$ in the problem domain and their error from $T_{FDM}$.
\begin{figure}[htbp!]
    \centering
    \includegraphics[width=0.9\textwidth]{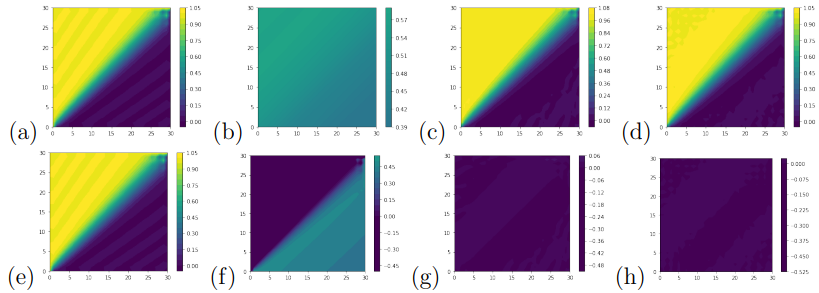}
    \caption{Temperature distribution at $Pe = 10, \; h = \tfrac{1}{30}$ with (a) FDM solution $T_{FDM}$ (b) PINN’s prediction of temperature with $\lambda_{DE} : \lambda_{DBC} = 1 : 1 \; \hat{T}_0$ (c) PINN’s prediction of temperature with $\lambda_{DE} : \lambda_{DBC} = Pe^{-1}h^2 : 1 \; \hat{T}_{NM}$ (d) PINN’s prediction of temperature with $\lambda_{DE} : \lambda_{DBC} = Pe^{-2}h^4 : 1 \; \hat{T}_{NM^2}$ (e) $T_{FDM}$ (f) Error between $\hat{T}_0$ and $T_{FDM}$ (g) Error between $\hat{T}_{NM}$ and $T_{FDM}$ (h) Error between $\hat{T}_{NM^2}$ and $T_{FDM}$.}
\end{figure}

The following figures show results of PINN with h = $\frac{1}{50}$ FIG.13 shows distribution of $\hat{T}_0$, $\hat{T}_{NM}$ and $\hat{T}_{NM^2}$ in the problem domain and their error from $T_{FDM}$.
\begin{figure}[htbp!]
    \centering
    \includegraphics[width=0.9\textwidth]{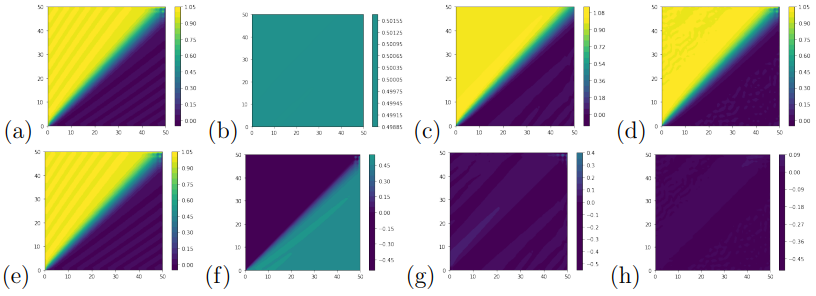}
    \caption{Temperature distribution at $Pe = 10, \; h = \tfrac{1}{50}$ with (a) FDM solution $T_{FDM}$ (b) PINN’s prediction of temperature with $\lambda_{DE} : \lambda_{DBC} = 1 : 1 \; \hat{T}_0$ (c) PINN’s prediction of temperature with $\lambda_{DE} : \lambda_{DBC} = Pe^{-1}h^2 : 1 \; \hat{T}_{NM}$ (d) PINN’s prediction of temperature with $\lambda_{DE} : \lambda_{DBC} = Pe^{-2}h^4 : 1 \; \hat{T}_{NM^2}$ (e) $T_{FDM}$ (f) Error between $\hat{T}_0$ and $T_{FDM}$ (g) Error between $\hat{T}_{NM}$ and $T_{FDM}$ (h) Error between $\hat{T}_{NM^2}$ and $T_{FDM}$.}
\end{figure}

\paragraph{Solution obtain by PINNs at Pe = 100}:

FIG.14 shows solutions obtained from PINNs and benchmark solution obtained from finite difference method $T_{FDM}$ at x = 0.5 when Pe = 100. FIG.14(a) shows that $\hat{T}_{NM}$ and $\hat{T}_{NM^2}$ agree with benchmark solution when h = $\frac{1}{10}$; FIG.14(b) shows that $\hat{T}_{NM}$ and $\hat{T}_{NM^2}$ agree with benchmark solution when h = $\frac{1}{30}$; and and FIG.14(c) shows that $\hat{T}_{NM}$ and $\hat{T}_{NM^2}$ agree with benchmark solution when h = $\frac{1}{50}$.

Corresponding mean square error(MSE) can be found in Table IV. The efficacy and accuracy of PINN trained with eq(49), eq(50) are thus demonstrated in this test problem.
\begin{figure}[htbp!]
    \centering
    % 第一張子圖
    \begin{subfigure}{0.32\textwidth}
        \centering
        \includegraphics[width=\textwidth]{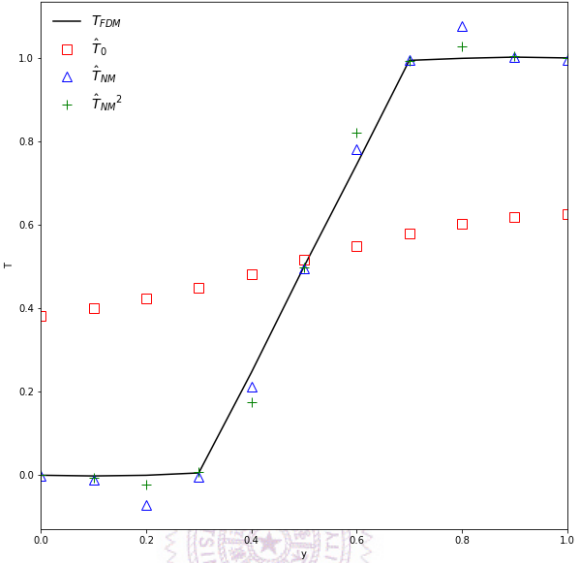}
        \caption{}
    \end{subfigure}
    % 第二張子圖
    \begin{subfigure}{0.32\textwidth}
        \centering
        \includegraphics[width=\textwidth]{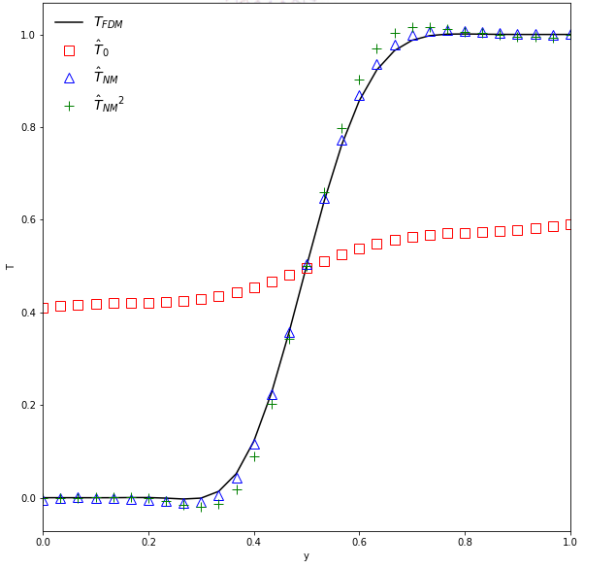}
        \caption{}
    \end{subfigure}
    % 第三張子圖
    \begin{subfigure}{0.32\textwidth}
        \centering
        \includegraphics[width=\textwidth]{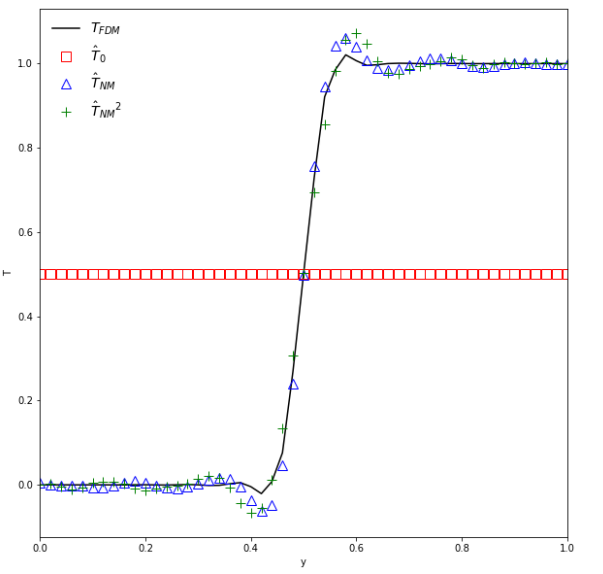}
        \caption{}
    \end{subfigure}

    \caption{  PINN’s prediction of temperature distribution at x = 0.5 at (a) $\frac{1}{10}$ (b) $\frac{1}{30}$ (c) $\frac{1}{50}$ }
\end{figure}

\begin{table}[h]
\centering
\renewcommand{\arraystretch}{1.2} % 行距微調
\begin{tabular}{c|c|c|c}
\hline
($\times 10^{-3}$) & $\hat{T}_0$ & $\hat{T}_{NM}$ & $\hat{T}_{NM^2}$ \\
Loss weight & eq(48) & eq(50) & eq(49) \\
\hline
$h = \tfrac{1}{10}$ & {\color{red} 135.1} & 5.523 & 5.169 \\
$h = \tfrac{1}{30}$ & {\color{red} 229.6} & 1.365 & 0.901 \\
$h = \tfrac{1}{50}$ & {\color{red} 236.0} & 0.724 & 0.709 \\
\hline
\end{tabular}
\caption{Mean square error (MSE) of solutions obtained from \newline convection-and-diffusion PINN at $Pe = 100$.}
\label{tab:conv-diff-MSE-Pe100}
\end{table}

The following figures show results of PINN with h = $\frac{1}{10}$. FIG. 15 shows distribution of $\hat{T}_0$, $\hat{T}_{NM}$ and $\hat{T}_{NM^2}$ in the problem domain and their error from $T_{FDM}$.
\begin{figure}[htbp!]
    \centering
    \includegraphics[width=0.9\textwidth]{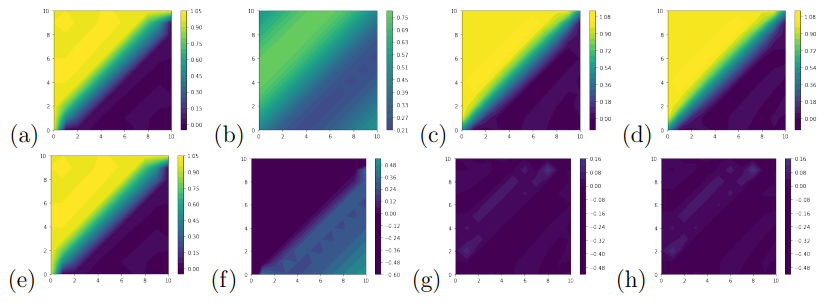}
    \caption{Temperature distribution at $Pe=100$, $h=\tfrac{1}{10}$ with (a) FDM solution $T_{FDM}$ (b) PINN’s prediction of temperature with $\lambda_{DE}:\lambda_{DBC}=1:1$ $\hat{T}_0$ (c) PINN’s prediction of temperature with $\lambda_{DE}:\lambda_{DBC}=Pe^{-1}h^2:1$ $\hat{T}_{NM}$ (d) PINN’s prediction of temperature with $\lambda_{DE}:\lambda_{DBC}=Pe^{-2}h^4:1$ $\hat{T}_{NM^2}$ (e) $T_{FDM}$ (f) Error between $\hat{T}_0$ and $T_{FDM}$ (g) Error between $\hat{T}_{NM}$ and $T_{FDM}$ (h) Error between $\hat{T}_{NM^2}$ and $T_{FDM}$.}
\end{figure}

The following figures show results of PINN with h = $\frac{1}{30}$. FIG.16 shows distribution of $\hat{T}_0$, $\hat{T}_{NM}$ and $\hat{T}_{NM^2}$ in the problem domain and their error from $T_{FDM}$.

\begin{figure}[htbp!]
    \centering
    \includegraphics[width=0.9\textwidth]{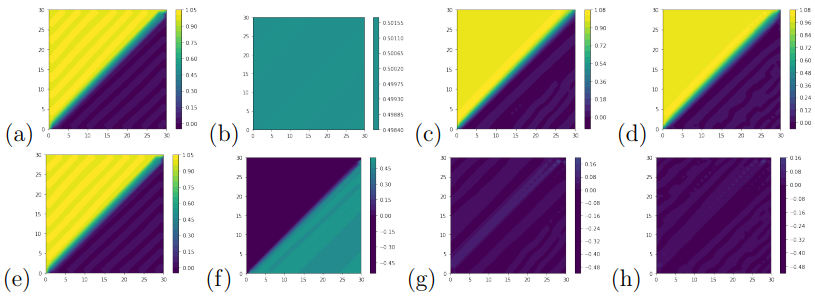}
    \caption{Temperature distribution at $Pe=100$, $h=\tfrac{1}{30}$ with (a) FDM solution $T_{FDM}$ (b) PINN’s prediction of temperature with $\lambda_{DE}:\lambda_{DBC}=1:1$ $\hat{T}_0$ (c) PINN’s prediction of temperature with $\lambda_{DE}:\lambda_{DBC}=Pe^{-1}h^2:1$ $\hat{T}_{NM}$ (d) PINN’s prediction of temperature with $\lambda_{DE}:\lambda_{DBC}=Pe^{-2}h^4:1$ $\hat{T}_{NM^2}$ (e) $T_{FDM}$ (f) Error between $\hat{T}_0$ and $T_{FDM}$ (g) Error between $\hat{T}_{NM}$ and $T_{FDM}$ (h) Error between $\hat{T}_{NM^2}$ and $T_{FDM}$.}
\end{figure}

The following figures show results of PINN with h = $\frac{1}{50}$. FIG.17 shows distribution of $\hat{T}_0$, $\hat{T}_{NM}$ and $\hat{T}_{NM^2}$ in the problem domain and their error from $T_{FDM}$.
\begin{figure}[htbp!]
    \centering
    \includegraphics[width=0.9\textwidth]{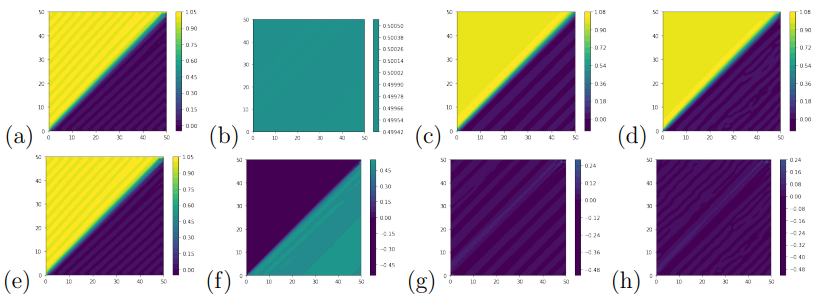}
    \caption{Temperature distribution at $Pe=100$, $h=\tfrac{1}{50}$ with (a) FDM solution $T_{FDM}$ (b) PINN’s prediction of temperature with $\lambda_{DE}:\lambda_{DBC}=1:1$ $\hat{T}_0$ (c) PINN’s prediction of temperature with $\lambda_{DE}:\lambda_{DBC}=Pe^{-1}h^2:1$ $\hat{T}_{NM}$ (d) PINN’s prediction of temperature with $\lambda_{DE}:\lambda_{DBC}=Pe^{-2}h^4:1$ $\hat{T}_{NM^2}$ (e) $T_{FDM}$ (f) Error between $\hat{T}_0$ and $T_{FDM}$ (g) Error between $\hat{T}_{NM}$ and $T_{FDM}$ (h) Error between $\hat{T}_{NM^2}$ and $T_{FDM}$.}
\end{figure}

\subsection{Lid-driven-cavity}
In this section, we compare the solutions obtained from PINNs trained with varying loss weight in Lid-driven-cavity PINN when Re = 10 and 100. A two-dimensional lid-driven cavity problem is governed by the steady state, two-dimensional incompressible Navier-Stokes equations and continuity, where Navier-Stokes equation in x-direction (x-momentum) is described as 
\begin{equation}
u \frac{\partial u}{\partial x} + v \frac{\partial u}{\partial y}
= -\frac{\partial p}{\partial x}
+ \frac{1}{Re}\left( \frac{\partial^2 u}{\partial x^2} + \frac{\partial^2 u}{\partial y^2} \right)
\end{equation}
Navier-Stokes equation in y-direction (y-momentum) is described as
\begin{equation}
u \frac{\partial v}{\partial x} + v \frac{\partial v}{\partial y}
= -\frac{\partial p}{\partial y}
+ \frac{1}{Re}\left( \frac{\partial^2 v}{\partial x^2} + \frac{\partial^2 v}{\partial y^2} \right)
\end{equation}
and continuity is described as
\begin{equation}
\frac{\partial u}{\partial x} + \frac{\partial v}{\partial y} = 0
\end{equation}
The domain boundary $\partial\Omega$ is defined with boundary condition of velocity in x-direction $g_u$ where
\begin{equation}
g_u(x,y) =
\begin{cases}
0, & \text{if } x = 0,1;\; y = 0 \\[6pt]
1, & \text{if } y = 1
\end{cases}
\end{equation}
,boundary condition of velocity in x-direction $g_v$ where
\begin{equation}
    g_v(x,y) = 0
\end{equation}
and boundary condition of pressure where
\begin{equation}
\frac{\partial p}{\partial n} = 0 \quad \text{if } (x,y) \in \partial \Omega
\end{equation}
\begin{figure}[htbp!]
    \centering
    \includegraphics[width=0.6\textwidth]{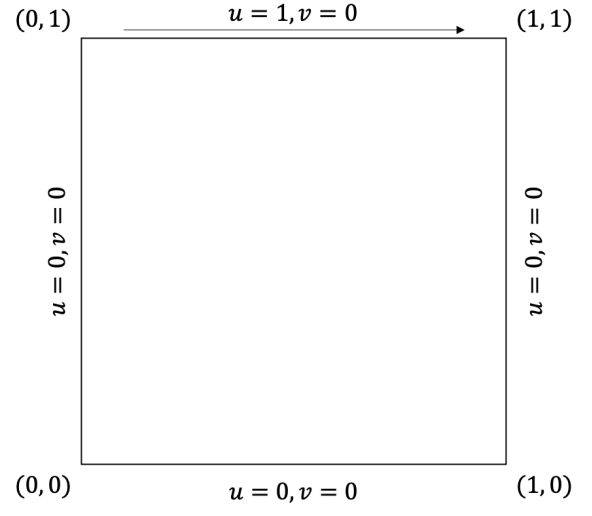}
    \caption{ Problem domain and boundary conditions of lid-driven-cavity problem}
\end{figure}

With the condition given in this problem, pressure gradient $p_x$ is unique while there exists infinite solution for pressure p. To evaluate PINN’s solution of pressure p, error between PINN’s solutions of pressure gradient $\hat{p}_x$ and benchmark for pressure gradient $\hat{p}_{xFDM}$ has has much more significance over error between PINNs’ solutions of pressure $\hat{p}$ and benchmark for pressure $p_{FDM}$ Thus, for FIG. 19, FIG. 20, and FIG. 21, besides the outputs of lid-driven-cavity PINN $\hat{u}$, $\hat{v}$ and $\hat{p}$ are plotted alongside, where $\hat{p}_x$ are processed from $\hat{p}$ with finite difference method(FDM) using central differencing scheme(CDS). Furthermore, velocity field is the focus in this problem, and thus we will focus on the velocity $\hat{u}$ obtain by PINN in this subsection.

\paragraph{Define loss function $\mathcal{L}$}:

A two-dimensional lid-driven-cavity problem is governed by eq(51), eq(52) and eq(53). The lid-driven-cavity PINN is trained with loss function $\mathcal{L}$ composed of Navier-Stokes equation in x-direction loss component $\mathcal{L}_{NSx}$, Navier-Stokes equation in y-direction loss component $\mathcal{L}_{NSy}$, continuity loss component $\mathcal{L}_{c}$ Dirichlet boundary condition loss component $\mathcal{L}_{DBC}$, and Neumann boundary condition loss component $\mathcal{L}_{NBC}$, each loss component is multiplied with a corresponding loss weight $\lambda$, which is defined as
\begin{equation}
\mathcal{L} = \lambda_{NS_x}\mathcal{L}_{NS_x}
+ \lambda_{NS_y}\mathcal{L}_{NS_y}
+ \lambda_{c}\mathcal{L}_{c}
+ \lambda_{DBC}\mathcal{L}_{DBC}
+ \lambda_{NBC}\mathcal{L}_{NBC}
\end{equation}
Loss components of conduction are defined by numerical differentiation(ND) with central differencing scheme(CDS), when compute with MSE, Navier-Stokes equation in x-direction loss component $\mathcal{L}_{NSx}$ is defined as
\begin{equation}
\begin{split}
\left\|
\left(u\frac{\partial u}{\partial x} + v\frac{\partial u}{\partial y}\right)
- \left(-\frac{\partial p}{\partial x} 
+ \frac{1}{Re}\left(\frac{\partial^2 u}{\partial x^2} + \frac{\partial^2 u}{\partial y^2}\right)\right)
\right\|_{\Omega}^2
& \\[6pt]
= \sum_{i=1}^{N-2}\sum_{j=1}^{N-2}
\Bigg\{
    u_{i,j}\cdot \frac{u_{i+1,j}-u_{i-1,j}}{2h}
  + v_{i,j}\cdot \frac{u_{i,j+1}-u_{i,j-1}}{2h}
  - \frac{p_{i+1,j}-p_{i-1,j}}{2h} \\
\qquad
  + \frac{1}{Re}\left(
      \frac{u_{i+1,j}+u_{i-1,j}+u_{i,j+1}+u_{i,j-1}-4u_{i,j}}{h^2}
    \right)
\Bigg\}^2
\end{split}
\end{equation}
,Navier-Stokes equation in y-direction loss component 
\begin{equation}
\begin{split}
\left\|
\left(u\frac{\partial v}{\partial x} + v\frac{\partial v}{\partial y}\right)
- \left(-\frac{\partial p}{\partial y} 
+ \frac{1}{Re}\left(\frac{\partial^2 v}{\partial x^2} + \frac{\partial^2 v}{\partial y^2}\right)\right)
\right\|_{\Omega}^2
& \\[6pt]
= \sum_{i=1}^{N-2}\sum_{j=1}^{N-2}
\Bigg\{
    u_{i,j}\cdot \frac{v_{i+1,j}-v_{i-1,j}}{2h}
  + v_{i,j}\cdot \frac{v_{i,j+1}-v_{i,j-1}}{2h}
  - \frac{p_{i,j+1}-p_{i,j-1}}{2h} \\
\qquad
  + \frac{1}{Re}\left(
      \frac{v_{i+1,j}+v_{i-1,j}+v_{i,j+1}+v_{i,j-1}-4v_{i,j}}{h^2}
    \right)
\Bigg\}^2
\end{split}
\end{equation}
,Navier-Stokes equation in y-direction loss component $\mathcal{L}_{NSy}$ is defined as
, continuity loss component $\mathcal{L}_c$ is defined as
\begin{equation}
\mathcal{L}_{c} =
\frac{\lVert \nabla \cdot \mathbf{u} \rVert_{\Omega}^{2}}{|\Omega|}
\end{equation}
,velocity boundary condition $g_u$ (eq(54)) and $g_v$ (eq(55)) is written into Dirichlet-boundary-condition loss component $\mathcal{L}_{DBC}$, where it is defined as
\begin{equation}
\mathcal{L}_{DBC} =
\frac{\lVert (u-g_u) + (v-g_v) \rVert_{\partial \Omega}^{2}}{|\partial \Omega|}
\end{equation}
where
\begin{equation}
\begin{split}
\left\|(u - g_u) + (v - g_v)\right\|_{\partial \Omega}^2
&= \sum_{i=0}^{N-2} \Big[ (u_{i,0} - g_{u,i,0})^2 + (v_{i,0} - g_{v,i,0})^2 \Big] \\
&\quad + \sum_{i=0}^{N-2} \Big[ (u_{i,N-1} - g_{u,i,N-1})^2 + (v_{i,N-1} - g_{v,i,N-1})^2 \Big] \\
&\quad + \sum_{j=0}^{N-2} \Big[ (u_{0,j} - g_{u,0,j})^2 + (v_{0,j} - g_{v,0,j})^2 \Big] \\
&\quad + \sum_{j=0}^{N-2} \Big[ (u_{N-1,j} - g_{u,N-1,j})^2 + (v_{N-1,j} - g_{v,N-1,j})^2 \Big]
\end{split}
\end{equation}
, pressure boundary condition $g_u$ (eq(54)) and $g_v$ (eq(55)) is written into Dirichlet-boundary-condition loss component $\mathcal{L}_{DBC}$, where it is defined as
\begin{equation}
\mathcal{L}_{NBC} = 
\frac{\left\| \frac{\partial p}{\partial n} - g_p \right\|^2_{\partial \Omega}}{|\partial \Omega|}
\end{equation}
where
\begin{equation}
\begin{aligned}
\left\| \frac{\partial p}{\partial n} - g_p \right\|^2_{\partial \Omega} 
&= \sum_{i=0}^{N-2} \left( \frac{p_{i+1,j} - p_{i-1,j}}{2h} \right)^2
+ \sum_{i=0}^{N-2} \left( \frac{p_{i+1,j} - p_{i-1,j}}{2h} \right)^2 \\
&\quad + \sum_{j=0}^{N-2} \left( \frac{p_{i,j+1} - p_{i,j-1}}{2h} \right)^2
+ \sum_{j=0}^{N-2} \left( \frac{p_{i,j+1} - p_{i,j-1}}{2h} \right)^2
\end{aligned}
\end{equation}
\paragraph{Dimensional Analysis}:

Two-dimensional lid-driven-cavity problem is governed by eq(51), eq(52) and eq(53). The lid-driven-cavity PINN is trained with loss function composed of Navier-Stokes equation in x-direction loss component $\mathcal{L}_{NSx}$ Navier-Stokes equation in y-direction loss component $\mathcal{L}_{NSy}$, continuity loss component $\mathcal{L}_c$, Dirichlet boundary condition loss component $\mathcal{L}_{DBC}$ and Neumann boundary condition loss component $\mathcal{L}_{NBC}$ each loss component is multiplied with a corresponding loss weight $\lambda$(eq(57)). To determine the proper ratio between each loss function , analyze on order of magnitude of loss component is considered. Order of magnitude of u momentum (Navier-Stokes equation in x direction) loss component is
\begin{equation}
\begin{aligned}
&\mathcal{L}_{NS_x} \sim \left(\frac{1}{Re} \cdot \frac{u_{i+1,j}+u_{i-1,j}+u_{i,j+1}+u_{i,j-1}-4u_{i,j}}{h^2}\right)^2
\sim \frac{[U]^2}{[Re]^2 [h]^4}
\end{aligned}
\end{equation}
Order of magnitude of v momentum (Navier-Stokes equation in y direction) loss component is 
\begin{equation}
\begin{aligned}
&\mathcal{L}_{NS_y}\sim \left(\frac{1}{Re} \cdot \frac{v_{i+1,j}+v_{i-1,j}+v_{i,j+1}+v_{i,j-1}-4v_{i,j}}{h^2}\right)^2
\sim \frac{[V]^2}{[Re]^2 [h]^4}
\end{aligned}
\end{equation}
Order of magnitude of continuity loss component is
\begin{equation}
\mathcal{L}_c = \frac{\|\nabla \cdot \mathbf{u}\|_{\Omega}^2}{|\Omega|}
\sim \frac{[U]^2}{[h]^2}
\end{equation}
Order of magnitude of Dirichlet boundary condition is
\begin{equation}
\mathcal{L}_{DBC} = \frac{\|(u-g_u) + (v-g_v)\|_{\partial \Omega}^2}{|\partial \Omega|}
\sim [U]^2
\end{equation}
Order of magnitude of Neumann boundary condition is
\begin{equation}
\mathcal{L}_{NBC} = 
\frac{\left\| \tfrac{\partial p}{\partial n} \right\|^2_{\partial\Omega}}
{|\partial\Omega|}
\sim \mathcal{L}_{NS_x}
\sim \frac{[U]^2}{[Re]^2 [h]^4}
\end{equation}
\paragraph{Investigation of Different Ratios of Loss Weight for Physics-Informed Neural Networks}:

Loss weights $\lambda_s$ in eq(57) control the contribution of each different component, we investigate three different weighting schemes for the loss components in Physics-Informed Neural Networks (PINNs). The three schemes are: when the loss weights are given the same value, when the ratio of loss weights is determined from the analysis of the order of magnitude, and when the square root of the ratio is employed as a relaxation factor. The first scheme represents the most commonly used approach for setting loss weights in PINNs, where
\begin{equation}
\lambda_{NS_x} : \lambda_{NS_y} : \lambda_c : \lambda_{DBC} : \lambda_{NBC} 
= 1 : 1 : 1 : 1 : 1
\end{equation}
Solutions obtained from PINNs trained ratio set with the first scheme will be denoted with a subscript ”0”, i.e, $\hat{u}_0$, $\hat{v}_0$ and $\hat{p}_0$.

The second scheme aims to balance the order of magnitude of the loss components by determining the loss weights based on the magnitude of quantifiable terms(eq(65), eq(66), eq(67), eq(68), eq(69)) within the loss components, where

\begin{equation}
\begin{aligned}
[\lambda_{NS_x}\mathcal{L}_{NS_x}]
\sim [\lambda_{NS_y}\mathcal{L}_{NS_y}]
\sim [\lambda_c\mathcal{L}_c]
\sim [\lambda_{DBC}\mathcal{L}_{DBC}]
\sim [\lambda_{NBC}\mathcal{L}_{NBC}] \\[6pt]
\lambda_{NS_x} : \lambda_{NS_y} : \lambda_c : \lambda_{DBC} : \lambda_{NBC}
\approx [Re]^2 h^4 : [Re]^2 h^4 : h^2 : 1 : [Re]^2 h^4
\end{aligned}
\end{equation}
Solutions obtained from PINNs trained ratio set with the second scheme will be denoted with a subscript ”$NM^2$", i.e, $\hat{u}_{NM^2}$, $\hat{v}_{NM^2}$ and $\hat{p}_{NM^2}$.

The third scheme introduces a relaxed version of the second scheme, taking into consideration that the magnitude of unquantifiable terms tends to change alongside the quantifiable terms. As a result, some relaxation is applied to the determined ratio, with the square root being used in this research as a form of relaxation, where
\begin{equation}
\lambda_{NS_x} : \lambda_{NS_y} : \lambda_c : \lambda_{DBC} : \lambda_{NBC} 
= Re \cdot h^2 : Re \cdot h^2 : h : 1 : Re \cdot h^2
\end{equation}
Solutions obtained from PINNs trained ratio set with the third scheme will be
denoted with a subscript ”NM”, i.e, $\hat{u}_{NM}$, $\hat{v}_{NM}$ and $\hat{p}_{NM}$.

\paragraph{Solution obtain by PINNs at Re = 100}:

FIG. 19, FIG. 20 and FIG. 21 show solutions obtained from PINNs and benchmark solution obtained from finite difference method when Re = 100. FIG. 19 shows that $\hat{u}_{0}$, $\hat{u}_{NM}$, and $\hat{u}_{NM^2}$ agree with benchmark solution when h = $\frac{1}{10}$. FIG. 20 shows that $\hat{u}_{0}$, $\hat{u}_{NM}$, and $\hat{u}_{NM^2}$ agree with benchmark solution when h = $\frac{1}{30}$; and FIG. 19(c) shows that $\hat{u}_{0}$, $\hat{u}_{NM}$ agree with benchmark solution when h = $\frac{1}{50}$.

Corresponding mean square error(MSE) can be found in Table V. The efficacy and accuracy of PINN trained with the first(eq(70)) and the third (eq(72)) weighting scheme are thus demonstrated in this test problem.
\begin{figure}[htbp!]
    \centering
    % 第一張子圖
    \begin{subfigure}{0.32\textwidth}
        \centering
        \includegraphics[width=\textwidth]{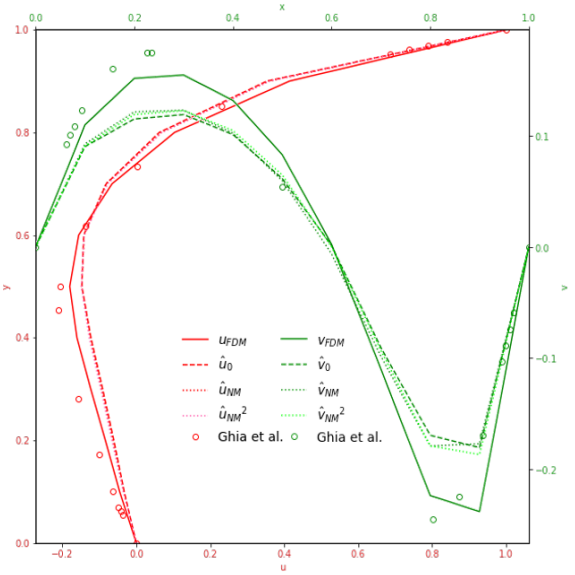}
        \caption{}
    \end{subfigure}
    % 第二張子圖
    \begin{subfigure}{0.32\textwidth}
        \centering
        \includegraphics[width=\textwidth]{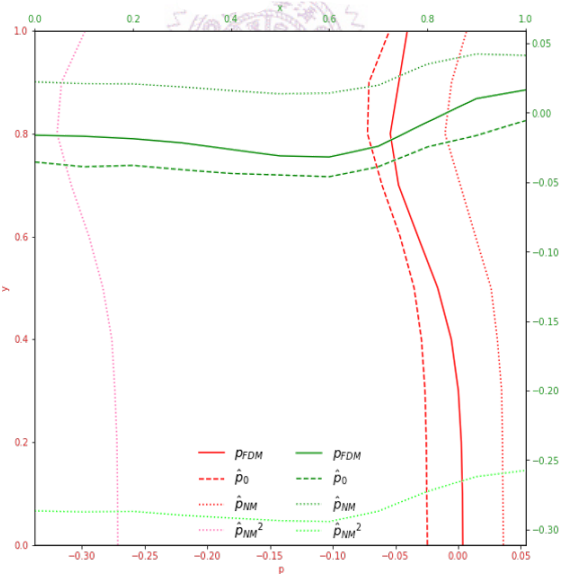}
        \caption{}
    \end{subfigure}
    % 第三張子圖
    \begin{subfigure}{0.32\textwidth}
        \centering
        \includegraphics[width=\textwidth]{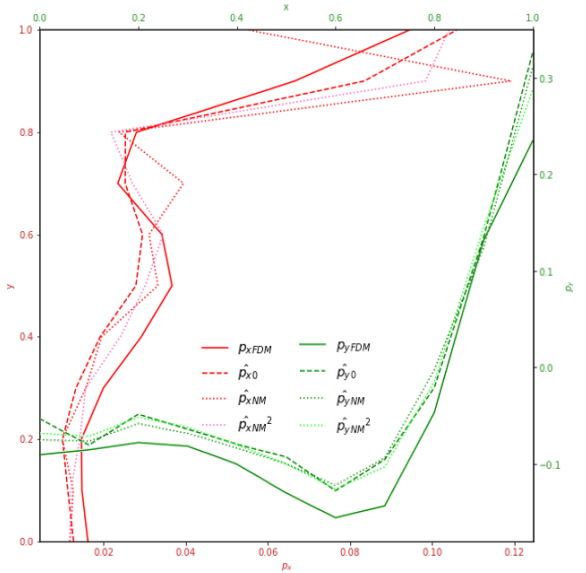}
        \caption{}
    \end{subfigure}
    \caption{PINN's prediction of (a) velocity $u(v)$ (b) pressure $p$ (c) pressure gradient $p_x(p_y)$ distribution at $x=0.5 (y=0.5)$ at $h=\tfrac{1}{10}$, Re=100.}

\end{figure}
\begin{figure}[htbp!]
    \centering
    % 第一張子圖
    \begin{subfigure}{0.32\textwidth}
        \centering
        \includegraphics[width=\textwidth]{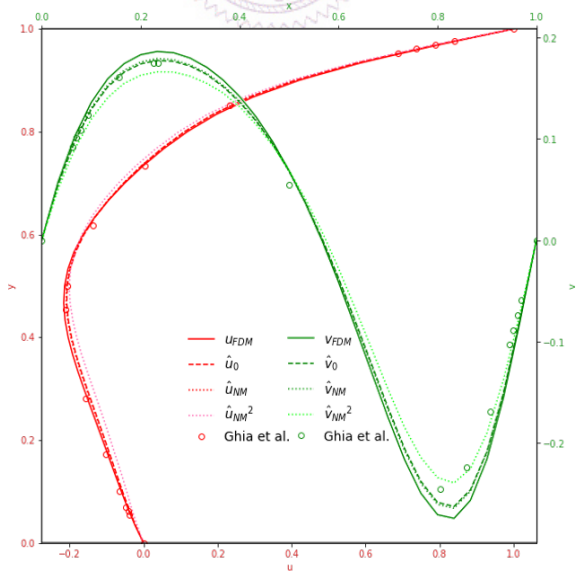}
        \caption{}
    \end{subfigure}
    % 第二張子圖
    \begin{subfigure}{0.32\textwidth}
        \centering
        \includegraphics[width=\textwidth]{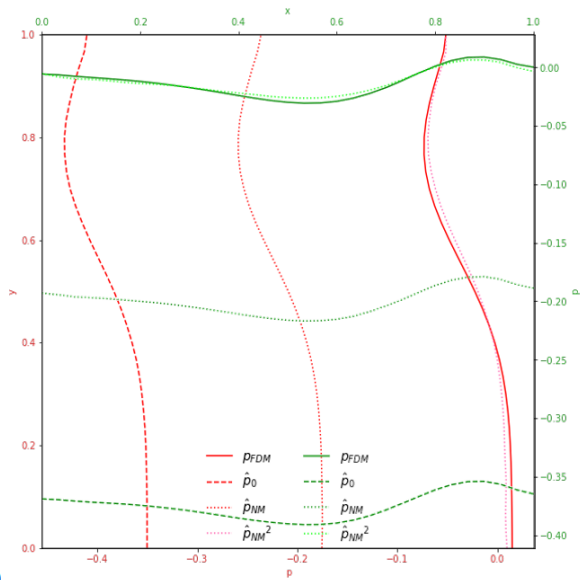}
        \caption{}
    \end{subfigure}
    % 第三張子圖
    \begin{subfigure}{0.32\textwidth}
        \centering
        \includegraphics[width=\textwidth]{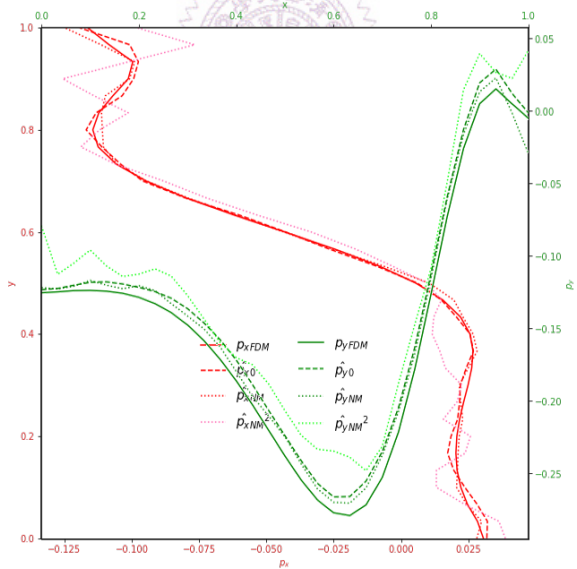}
        \caption{}
    \end{subfigure}
    \caption{PINN's prediction of (a) velocity $u(v)$ (b) pressure $p$ (c) pressure gradient $p_x(p_y)$ distribution at $x=0.5 (y=0.5)$ at $h=\tfrac{1}{30}$, Re=100.}

\end{figure}
\begin{figure}[htbp!]
    \centering
    % 第一張子圖
    \begin{subfigure}{0.32\textwidth}
        \centering
        \includegraphics[width=\textwidth]{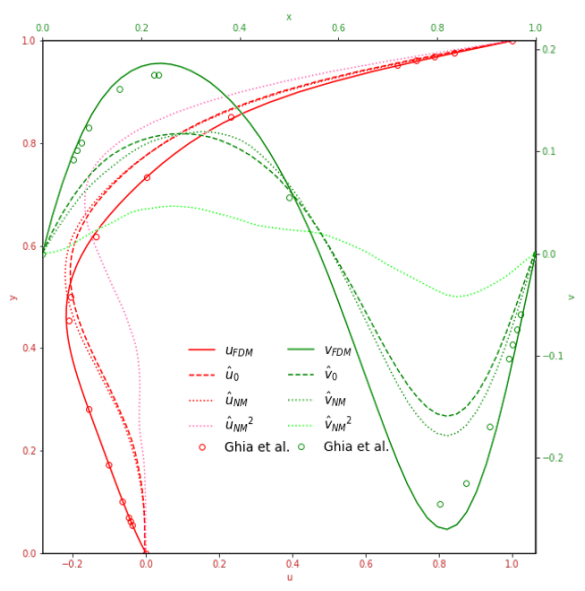}
        \caption{}
    \end{subfigure}
    % 第二張子圖
    \begin{subfigure}{0.32\textwidth}
        \centering
        \includegraphics[width=\textwidth]{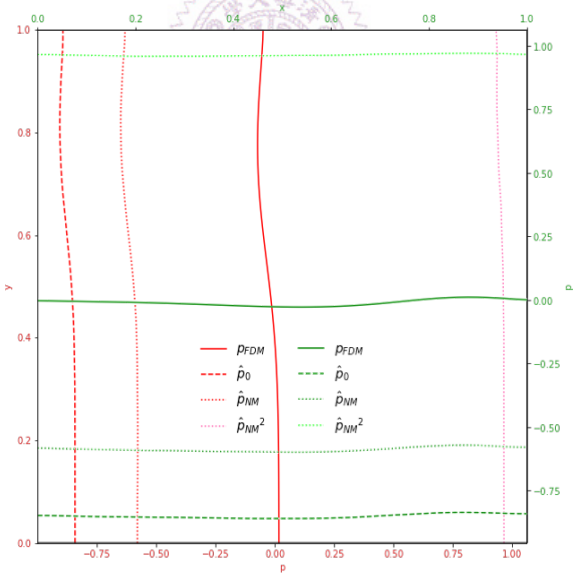}
        \caption{}
    \end{subfigure}
    % 第三張子圖
    \begin{subfigure}{0.32\textwidth}
        \centering
        \includegraphics[width=\textwidth]{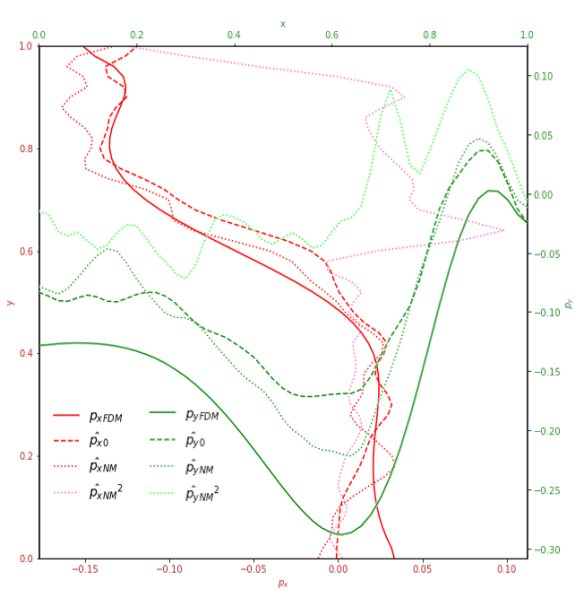}
        \caption{}
    \end{subfigure}
    \caption{PINN's prediction of (a) velocity $u(v)$ (b) pressure $p$ (c) pressure gradient $p_x(p_y)$ distribution at $x=0.5 (y=0.5)$ at $h=\tfrac{1}{50}$, Re=100.}

\end{figure}

\begin{table}[h]
\centering
\renewcommand{\arraystretch}{1.2} % 調整行距
\begin{tabular}{c|c|c|c}
\hline
($\times 10^{-3}$) 
& $\sqrt{\hat{u}_0^2 + \hat{v}_0^2}$ 
& $\sqrt{\hat{u}_{NM}^2 + \hat{v}_{NM}^2}$ 
& $\sqrt{\hat{u}_{NM^2}^2 + \hat{v}_{NM^2}^2}$ \\
Loss weight 
& eq(70) 
& eq(72) 
& eq(71) \\
\hline
$h = \tfrac{1}{10}$ & 4.778 & 4.735 & 4.701 \\
$h = \tfrac{1}{30}$ & 0.567 & 0.553 & 0.771 \\
$h = \tfrac{1}{50}$ & 2.466 & 2.088 & {\color{red} 11.07} \\
\hline
\end{tabular}
\caption{Mean square error (MSE) of solutions obtained \newline from lid-driven-cavity PINN at $Re = 100$.}
\label{tab:lid-driven-MSE-Re100}
\end{table}

The following figures show results of PINN with h = $\frac{1}{10}$. FIG. 22 shows distribution of $\hat{u}_{0}$, $\hat{u}_{NM}$, and $\hat{u}_{NM^2}$ in the problem domain and their error from $u_{FDM}$.
\begin{figure}[htbp!]
    \centering
    \includegraphics[width=0.8\textwidth]{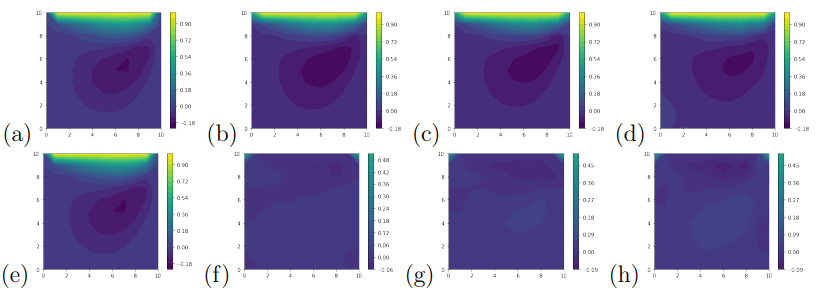}
    \caption{x-component of velocity at $Re=100$, $h=\tfrac{1}{10}$ with (a) FDM solution $u_{FDM}$ 
    (b) PINN's prediction of x-component of velocity with $\lambda_{NS_x} : \lambda_{NS_y} : \lambda_c : \lambda_{DBC} : \lambda_{NBC} = 1:1:1:1:1:1 \; \hat{u}_0$ 
    (c) PINN's prediction of x-component of velocity with $\lambda_{NS_x} : \lambda_{NS_y} : \lambda_c : \lambda_{DBC} : \lambda_{NBC} = Re \cdot h^2 : Re \cdot h^2 : h : 1 : Re h^2 \; \hat{u}_{NM}$ 
    (d) PINN's prediction of x-component of velocity with $\lambda_{NS_x} : \lambda_{NS_y} : \lambda_c : \lambda_{DBC} : \lambda_{NBC} = Re^2 h^4 : Re^2 h^4 : h^2 : 1 : Re^2 h^4 \; \hat{u}_{NM^2}$ 
    (e) $u_{FDM}$ (f) Error between $\hat{u}_0$ and $u_{FDM}$ (g) Error between $\hat{u}_{NM}$ and $u_{FDM}$ (h) Error between $\hat{u}_{NM^2}$ and $u_{FDM}$.}
\end{figure}

The following figures show results of PINN with h = $\frac{1}{30}$. FIG. 23 shows distribution of $\hat{u}_{0}$, $\hat{u}_{NM}$, and $\hat{u}_{NM^2}$ in the problem domain and their error from $u_{FDM}$.
\begin{figure}[htbp!]
    \centering
    \includegraphics[width=0.9\textwidth]{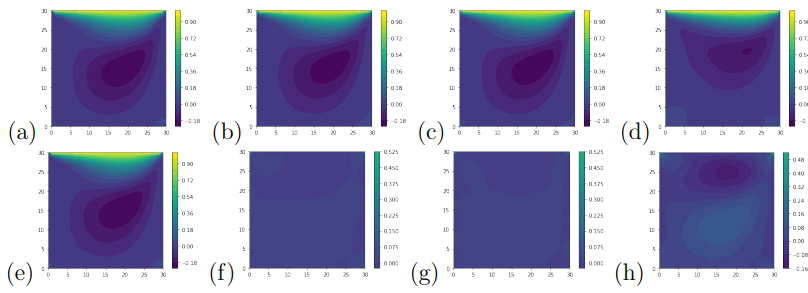}
    \caption{%
    x-component of velocity at $Re = 100, \, h = \tfrac{1}{30}$ with 
    (a) FDM solution $u_{FDM}$ 
    (b) PINN’s prediction of x-component of velocity with 
    $\lambda_{NS_x} : \lambda_{NS_y} : \lambda_c : \lambda_{DBC} : \lambda_{NBC} = 1 : 1 : 1 : 1 : 1  \hat{u}_0$ 
    (c) PINN’s prediction of x-component of velocity with 
    $\lambda_{NS_x} : \lambda_{NS_y} : \lambda_c : \lambda_{DBC} : \lambda_{NBC} = Re \cdot h^2 : Re \cdot h^2 : h : 1 : Re h^2 \, \hat{u}_{NM}$ 
    (d) PINN’s prediction of x-component of velocity with 
    $\lambda_{NS_x} : \lambda_{NS_y} : \lambda_c : \lambda_{DBC} : \lambda_{NBC} = Re^2 h^4 : Re^2 h^4 : h^2 : 1 : Re^2 h^4 \, \hat{u}_{NM^2}$ 
    (e) $u_{FDM}$ 
    (f) Error between $\hat{u}_0$ and $u_{FDM}$ 
    (g) Error between $\hat{u}_{NM}$ and $u_{FDM}$ 
    (h) Error between $\hat{u}_{NM^2}$ and $u_{FDM}$.}
\end{figure}
The following figures show results of PINN with h = $\frac{1}{50}$. FIG. 24 shows distribution of $\hat{u}_{0}$, $\hat{u}_{NM}$, and $\hat{u}_{NM^2}$ in the problem domain and their error from $u_{FDM}$.
\begin{figure}[htbp!]
    \centering
    \includegraphics[width=0.9\textwidth]{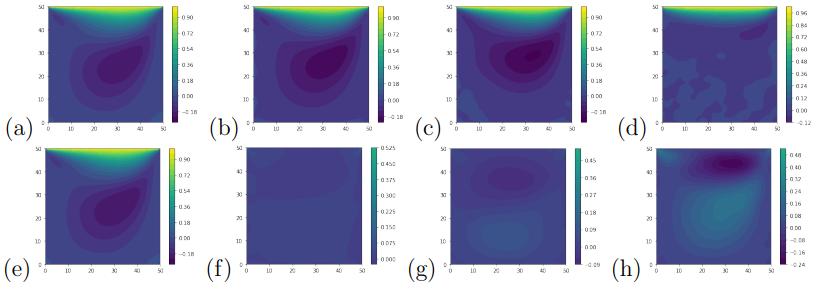}
    \caption{%
    x-component of velocity at $Re=100$, $h=\tfrac{1}{50}$ with 
    (a) FDM solution $u_{FDM}$ 
    (b) PINN’s prediction of x-component of velocity with $\lambda_{NS_x}:\lambda_{NS_y}:\lambda_c:\lambda_{DBC}:\lambda_{NBC}=1:1:1:1:1:\hat{u}_0$ 
    (c) PINN’s prediction of x-component of velocity with $\lambda_{NS_x}:\lambda_{NS_y}:\lambda_c:\lambda_{DBC}:\lambda_{NBC}=Re\cdot h^2:Re\cdot h^2:h:1:Re h^2 \hat{u}_{NM}$ 
    (d) PINN’s prediction of x-component of velocity with $\lambda_{NS_x}:\lambda_{NS_y}:\lambda_c:\lambda_{DBC}:\lambda_{NBC}=Re^2h^4:Re^2h^4:h^2:1:Re^2h^4 \hat{u}_{NM^2}$ 
    (e) $u_{FDM}$ 
    (f) Error between $\hat{u}_0$ and $u_{FDM}$ 
    (g) Error between $\hat{u}_{NM}$ and $u_{FDM}$ 
    (h) Error between $\hat{u}_{NM^2}$ and $u_{FDM}$.
    }
\end{figure}

\paragraph{Computation efficiency of C++ and python with PyTorch}:

Compiled languages such as C/C++ and FORTRAN translate directly to machine code and thus run efficiently, making them well-suited for numerical computation. In contrast, interpreted languages like Python execute via bytecode, resulting in lower efficiency and making them less ideal for high-performance numerical codes.  

Despite this drawback, Python is widely adopted for its simplicity and versatility, and frameworks such as TensorFlow and PyTorch are primarily developed in Python with back end optimizations tailored for it. This raises the question of whether PyTorch-based numerical code in C++ can outperform Python, motivating the comparative tests conducted in this study.  
\begin{table}[h]
\centering
\renewcommand{\arraystretch}{1.2}
\setlength{\tabcolsep}{8pt} % 調整欄間距
\begin{tabular}{l c c}
\hline
\textbf{Description of test cases} & \textbf{C++} & \textbf{Python} \\
\hline
(1) Solve Poisson equation on a $80\times80$ grid for 1000 times. & 0.0756 & 12.9560 \\
(2) Generate a $80\times80$ output by a ML model for 1000 times.  & 2.0863 & 0.1880 \\
(1)+(2) & 2.6631 & 14.9609 \\
\hline
\end{tabular}

\caption{Comparison of C++ and Python in execution time (second).}
\label{tab:python-time}
\end{table}

From the table above, we found that overall performance indicates that integrating learning-based method in python to numerical solver in C++ is a faster choice for integrating numerical solver and learning-based method, and therefore C++ is chosen as the language for integrating the two method.

\section{Conclusion}\label{sec:4}
In this thesis, we investigate loss-weighting strategies for Physics-Informed Neural Networks (PINNs). Two schemes are proposed: the first assigns weights based on the orders of magnitude of quantifiable loss terms, while the second also accounts for unquantifiable terms. These are compared with the commonly used equal-weight scheme across three benchmark problems: conduction, convection–diffusion, and lid-driven cavity.

Our results show that the second scheme achieves superior accuracy in the conduction and lid-driven cavity problems, underscoring the importance of incorporating both quantifiable and unquantifiable terms. For the convection–diffusion case, both schemes perform effectively, capturing the complex underlying physics.

An additional finding is that PINNs can solve equations that are unstable or unsolvable with traditional numerical methods, demonstrating their potential for tackling challenging problems in computational physics. Overall, our study highlights the significance of informed loss weighting in enhancing PINN performance and broadening their applicability.

\begin{acknowledgments}
Chao-An Lin would like to acknowledge support by Taiwan National Science and Technology Council under project No. NSTC 113-2221-E-007-129.
\end{acknowledgments}

\section*{Data Availability Statement}
The data supporting this study's findings are available from the corresponding author upon reasonable request.

\section*{Conflict of Interest}
The authors have no conflicts to disclose.

\bibliographystyle{aipnum4-2}  % 或 apsrev4-2，看你用哪個樣式
%aipnum4-2.bst 2019-01-14 (MD) hand-edited version of apsrev4-1.bst

\clearpage
    
\end{document}